\documentclass{article}

\usepackage{iclr_aims}

\newif\ifICLRversion
\ICLRversionfalse 

\newcommand{\ifICLR}[1]{\ifICLRversion #1\fi}
\newcommand{\ifOP}[1]{\ifICLRversion\else #1\fi}

\iclrfinalcopy
\usepackage{booktabs}
\usepackage{tcolorbox}
\tcbuselibrary{breakable}
\usepackage[utf8]{inputenc} 
\usepackage[T1]{fontenc}    
\usepackage{times}          
\usepackage{url}            
\usepackage{booktabs}       
\usepackage{amsfonts}       
\usepackage{nicefrac}       
\usepackage{microtype}      
\usepackage{lipsum}
\usepackage{graphicx}       
\usepackage{xcolor}         
\usepackage{array}          
\usepackage{amsmath}        
\usepackage{tikz}
\usetikzlibrary{positioning, shapes, arrows, fit, backgrounds, calc}
\usepackage{xcolor}
\usepackage{float}
\usepackage{placeins}       
\usepackage{eurosym}        
\usepackage{enumitem}       
\usepackage{makecell}       
\usepackage{fancyvrb}
\usepackage{listings}

\usepackage{array}          
\usepackage{tabularx}       
\usepackage{amsmath}        
\usepackage[bookmarks=true,bookmarksnumbered=true]{hyperref}  
\usepackage{bookmark}       

\graphicspath{{images/}}     

\title{Automating Forecasting Question Generation and Resolution for AI Evaluation

}

\ifOP{
\author{Nikos I. Bosse, Peter M\"uhlbacher, Jack Wildman, Lawrence Phillips, Dan Schwarz \\
FutureSearch \\
\texttt{\{nikos, peter, jack, lawrence, dan\}@futuresearch.ai}}
}
\ifICLR{\author{}}

\newcommand{\yes}{\textsc{yes}}
\newcommand{\no}{\textsc{no}}

\newcommand{\gptfive}{GPT-5}
\newcommand{\gptfivemini}{GPT-5 Mini}
\newcommand{\gptfivetwo}{GPT-5.2}
\newcommand{\geminithreepro}{Gemini 3 Pro}
\newcommand{\geminitwofivepro}{Gemini 2.5 Pro}
\newcommand{\geminitwofiveflash}{Gemini 2.5 Flash}
\newcommand{\claudehaiku}{Claude Haiku 4.5}
\newcommand{\claudeopus}{Opus 4.5}

\begin{document}

\maketitle
\thispagestyle{fancy}
\lhead{Published at ICLR 2026 Workshop on AI for Mechanism Design and Strategic Decision Making.}

\begin{abstract}

  Forecasting future events is highly valuable in decision-making and is a robust measure of general intelligence. As forecasting is probabilistic, developing and evaluating AI forecasters requires generating large numbers of diverse and difficult questions, and accurately resolving them. Previous efforts to automate this laborious work relied on recurring data sources (e.g., weather, stocks), limiting diversity and utility. In this work, we present a system for generating and resolving high-quality forecasting questions automatically and at scale using LLM-powered web research agents. We use this system to generate 1499 diverse, real-world forecasting questions, and to resolve them several months later. We estimate that our system produces verifiable, unambiguous questions approximately 96\% of the time, exceeding the rate of Metaculus, a leading human-curated forecasting platform. We also find that our system resolves questions at approximately 95\% accuracy. We verify that forecasting agents powered by more intelligent LLMs perform better on these questions (Brier score of 0.134 for Gemini 3 Pro, 0.149 for GPT-5, and 0.179 for Gemini 2.5 Flash). Finally, we demonstrate how our system can be leveraged to directly improve forecasting, by evaluating a question decomposition strategy on a generated question set, yielding a significant improvement in Brier scores (0.132 vs. 0.141).

\end{abstract}

\section{Introduction}

Recent work has demonstrated that large language models (LLMs) can be usefully applied to forecasting, approaching or matching human-level performance when equipped with appropriate tools and scaffolding \citep{halawi2024approaching, hsiehReasoningToolsHuman2024, schoeneggerWisdomSiliconCrowd2024}. This is an interesting development for at least two reasons. First, it implies that AI forecasters are poised to provide valuable insights into key real-world decisions. Second, forecasting is a robust and ungameable measure of general intelligence, and progress on forecasting tasks is an unmistakable sign of civilization's advancement towards Artificial General Intelligence (AGI). Thus, tracking progress in this domain is important both so that decision makers are appropriately calibrated with respect to capabilities they may soon use or consider using, and as an input into AGI timeline estimates.

However, obtaining the data necessary to evaluate AI forecasting systems presents a challenge. Forecasting is an especially sample-hungry capability: Unlike classification tasks where each example provides a definitive label, the ``gold label'' for a forecast is a probability distribution representing genuine uncertainty about the future. For any given question, however, we observe only a single draw from this distribution: the event either occurs or it does not. This dramatically increases sample complexity---distinguishing a well-calibrated 70\% forecast from a poorly-calibrated 60\% forecast requires observing outcomes across many questions \citep{bosseComparingTwoForecasters2023}. At the same time, forecasting questions are time-consuming to design correctly. Each question requires careful specification of resolution criteria and consideration of potential ambiguities to ensure objective resolution. Online prediction markets and forecasting platforms such as Polymarket, Kalshi, or \citet{metaculusForecastingPlatform} invest substantial human effort in question generation, yet even professionally curated questions frequently require annulment or face disputes over resolution.

The difficulty of creating and resolving high-quality forecasting questions at scale has constrained empirical research in this area. Existing benchmarks for evaluating AI forecasters, including the Metaculus AI Forecasting Benchmark Series \citep{metaculusAIBenchmarking2024}, ForecastBench \citep{karger2025forecastbench}, and Bench to the Future \citep{futuresearchBenchFuturePastcasting2025}, rely either on labor-intensive human curation or on automated generation from recurring data sources (e.g., economic indicators, sports statistics), which produces highly correlated or relatively trivial questions. Some studies consequently struggle with statistical power, limiting the conclusions that can be drawn \citep{schoeneggerWisdomSiliconCrowd2024}.

In this work, we make progress toward alleviating this data shortage by developing a system for generating and resolving high-quality forecasting questions at scale using LLMs. Since evaluation is our primary use case, the question sets our system generate should satisfy the following desiderata:
\begin{itemize}[nosep]
  \item \textbf{Unambiguously resolvable}: Questions should have clear, objective resolution criteria that can be verified by the system in practice.
  \item \textbf{Accurately resolved}: So long as a generated question is well formed, the system should resolve it accurately with high reliability.
  \item \textbf{Difficult}: Expected performance on each question should increase with the intelligence and effort applied to it, up to high levels of both. This ensures the benchmark remains useful as models improve, and reflects the nature of consequential real-world forecasting problems.
  \item \textbf{Consequential}: Questions should exercise forecasting capabilities that generalize to important use cases, rather than focusing on trivia or narrow domains.
  \item \textbf{Diverse}: The question set should span varied topics and reasoning approaches. This encourages general-purpose intelligence and research capabilities, and avoids losing statistical power through inter-question correlation.
\end{itemize}
A detailed taxonomy of desirable qualities for forecasting questions is provided in Appendix~\ref{sec:appendix-desirable-qualities}.

Our approach exploits a common asymmetry: in the case of question generation, as with many capabilities, verifying output quality is substantially easier than generating high-quality outputs from scratch. We employ an agentic workflow in which LLM agents using the live web propose candidate questions, which are then filtered by a series of targeted verifiers that assess resolvability, difficulty, and other criteria. To our knowledge, this is the first forecasting question generation system to leverage flexible web-research agents, enabling the production of timely, grounded questions that would be difficult to generate from static knowledge alone.

We evaluate our system by using it to generate a set of 1499 questions, and assessing these via multiple methodologies: through expert human ratings of question quality, by clustering questions to assess content topic and within-cluster diversity, by comparing annulment rates against the baseline of human-written Metaculus questions, and by verifying that more capable AI systems achieve better forecasting performance on our questions.

\section{Question Generation Pipeline}
\label{sec:question-generation-pipeline}

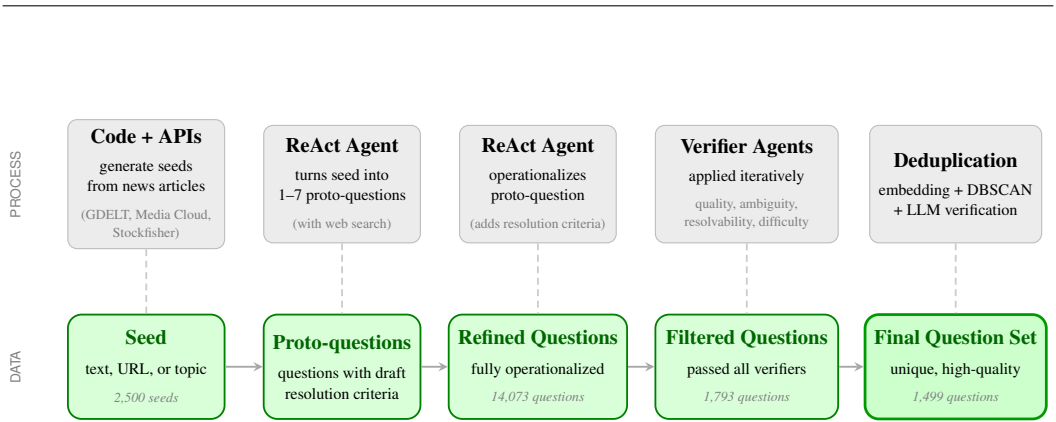
\begin{figure}[tbp]
  \centering
  \resizebox{\textwidth}{!}{%
    \begin{tikzpicture}[
  processbox/.style={
    rectangle, rounded corners=4pt,
    fill=gray!15, draw=gray!50, line width=0.5pt,
    minimum width=2.4cm, minimum height=1.8cm,
    font=\small, align=center, inner sep=4pt
  },
  databox/.style={
    rectangle, rounded corners=6pt,
    fill=green!15, draw=green!50!black, line width=0.8pt,
    minimum width=2.4cm, minimum height=1.6cm,
    font=\small, align=center, inner sep=4pt
  },
  finalbox/.style={
    rectangle, rounded corners=6pt,
    fill=green!20, draw=green!60!black, line width=1.2pt,
    minimum width=2.4cm, minimum height=1.6cm,
    font=\small, align=center, inner sep=4pt
  },
  arrow/.style={->, >=stealth, thick, gray!70},
  connector/.style={densely dashed, gray!40, line width=0.8pt}
]

\def\xsep{3.0cm}

\node[processbox] (p1) at (0,0) {\textbf{Code + APIs}\\[2pt]{\scriptsize generate seeds}\\[-1pt]{\scriptsize from news articles}\\[2pt]{\tiny\color{gray}(GDELT, Media Cloud,}\\[-2pt]{\tiny\color{gray}Stockfisher)}};

\node[processbox] (p2) at (\xsep,0) {\textbf{ReAct Agent}\\[2pt]{\scriptsize turns seed into}\\[-1pt]{\scriptsize 1--7 proto-questions}\\[2pt]{\tiny\color{gray}(with web search)}};

\node[processbox] (p3) at (2*\xsep,0) {\textbf{ReAct Agent}\\[2pt]{\scriptsize operationalizes}\\[-1pt]{\scriptsize proto-question}\\[2pt]{\tiny\color{gray}(adds resolution criteria)}};

\node[processbox, minimum width=2.8cm] (p4) at (3*\xsep+0.2cm,0) {\textbf{Verifier Agents}\\[2pt]{\scriptsize applied iteratively}\\[2pt]{\tiny\color{gray}quality, ambiguity,}\\[-2pt]{\tiny\color{gray}resolvability, difficulty}};

\node[processbox] (p5) at (4*\xsep+0.4cm,0) {\textbf{Deduplication}\\[2pt]{\scriptsize embedding + DBSCAN}\\[-1pt]{\scriptsize + LLM verification}};

\node[databox] (d1) at (0,-2.8cm) {\textbf{\color{green!40!black}Seed}\\[3pt]{\scriptsize text, URL, or topic}\\[2pt]{\tiny\itshape\color{gray}2,500 seeds}};

\node[databox] (d2) at (\xsep,-2.8cm) {\textbf{\color{green!40!black}Proto-questions}\\[3pt]{\scriptsize questions with draft}\\[-1pt]{\scriptsize resolution criteria}};

\node[databox] (d3) at (2*\xsep,-2.8cm) {\textbf{\color{green!40!black}Refined Questions}\\[3pt]{\scriptsize fully operationalized}\\[2pt]{\tiny\itshape\color{gray}14,073 questions}};

\node[databox, minimum width=2.8cm] (d4) at (3*\xsep+0.2cm,-2.8cm) {\textbf{\color{green!40!black}Filtered Questions}\\[3pt]{\scriptsize passed all verifiers}\\[2pt]{\tiny\itshape\color{gray}1,793 questions}};

\node[finalbox] (d5) at (4*\xsep+0.4cm,-2.8cm) {\textbf{\color{green!40!black}Final Question Set}\\[3pt]{\scriptsize unique, high-quality}\\[2pt]{\tiny\itshape\color{gray}1,499 questions}};

\draw[arrow] (d1) -- (d2);
\draw[arrow] (d2) -- (d3);
\draw[arrow] (d3) -- (d4);
\draw[arrow] (d4) -- (d5);

\draw[connector] (p1) -- (d1);
\draw[connector] (p2) -- (d2);
\draw[connector] (p3) -- (d3);
\draw[connector] (p4) -- (d4);
\draw[connector] (p5) -- (d5);

\node[rotate=90, anchor=south, font=\tiny\sffamily, gray] at (-1.8cm, 0) {PROCESS};
\node[rotate=90, anchor=south, font=\tiny\sffamily, gray] at (-1.8cm, -2.8cm) {DATA};

\end{tikzpicture}%
  }
  \caption{Question generation pipeline. Top row shows the processing steps; bottom row shows the data at each stage. Dashed lines connect each process to its output.}
  \label{fig:question-generation-pipeline}
\end{figure}

Our question-generation system consists of a multi-stage pipeline that iteratively calls LLM agents to suggest, refine, and critique forecasting questions (see Figure \ref{fig:question-generation-pipeline}).

\paragraph{Web agents.} ReAct-style agents \citep{yao2023reactsynergizingreasoningacting} that carry out web research are used throughout our pipeline and are fundamental to our approach. \ifOP{Specifically, we use Everyrow's ReAct implementation \citep{everyrow2026}, which is optimized for accurate web research at scale.}\ifICLR{Specifically, we developed a ReAct agent optimized for web research.} The agent's system prompt, toolkit, and time-management mechanism were tuned against Deep Research Bench \citep{futuresearch2025deepresearchbench}, a benchmark composed of challenging and realistic web-research tasks. All subsequent references to ReAct agents in this paper refer to this optimized implementation.

\paragraph{Question seeds.} The pipeline begins with a seed---an arbitrary piece of text intended as inspiration for the question's subject matter. Seeds serve to ground the system in specific real-world events and provide users with control over the domain of generated questions. Without such grounding, LLMs tend to propose topics that are either too broad to yield precise, resolvable questions or that drift from reality due to limitations in their detailed knowledge of current events. Additionally, asking LLMs to generate diverse seed ideas directly proves difficult in practice: their outputs tend toward repetition unless considerable effort is invested in prompting for variety. Suitable seed sources include news articles, company reports, or any other text describing current events or anticipated developments.

\paragraph{Proto-questions.} Based on the question seeds, we used a ReAct agent to propose a set of 1--7 forecasting proto-questions per seed. A proto-question is a question about the future whose general thrust is clear, but whose formulation is not precise enough to be objectively resolvable. Agentic web search is key for performance in this step---earlier experiments revealed that, without additional research, LLMs are prone to generating plausible-but-invalid questions (for example, "Will a certain space mission (that was, in fact, moved to 2026) happen until the end of 2025?") and to hallucinating resolution sources.
The prompt for the agent can be found in the Appendix \ref{sec:appendix-prompt-seed-to-proto-question}.

\paragraph{Refined forecasting questions.} Based on every proto-question, we asked another ReAct agent to refine the question, adding precise, objective resolution criteria. The prompt for the agent can be found in the Appendix \ref{sec:appendix-prompt-proto-question-to-refined-question}.

\paragraph{Question verifiers.} Next, we applied a set of verifiers to each refined forecasting question to filter for high-quality candidates. Each verifier is a separate ReAct agent, tasked with evaluating a specific aspect of the question:
\begin{enumerate}
  \item Overall question quality: Is this a good forecasting question in the sense that it poses a non-trivial forecasting task where higher research efforts and better judgement should on average lead to better performance? (See prompt in Appendix \ref{sec:appendix-prompt-assess-question-is-good}).
  \item Question ambiguity: Will the question be unambiguously resolvable? (See prompt in Appendix \ref{sec:appendix-prompt-assess-ambiguity}).
  \item Question resolvability: Will it be possible to resolve the question automatically using another ReAct agent? (See prompt in Appendix \ref{sec:appendix-prompt-assess-automated-resolvability}).
  \item Forecast: A simple forecaster that produces a probability forecast for the question to check for triviality. (See prompt in Appendix \ref{sec:appendix-prompt-assess-question-via-forecasting}).
\end{enumerate}

The first three verifiers output categorical ratings. The quality and ambiguity verifiers use a four-point scale of ``bad'', ``meh'', ``good'', or ``great''; questions must receive ``great'' on both to pass filtering. The resolvability verifier outputs ``very certainly no'', ``probably no'', ``probably yes'', or ``very certainly yes''; questions must receive ``very certainly yes'' to pass. The forecast verifier outputs a probability between 0 and 100; we initially intended to filter questions with forecasts below 2\% or above 98\% as trivial, but retained them since they constituted only a small fraction of the total.

\paragraph{Question deduplication.} To ensure question uniqueness, we applied a simple automated deduplication procedure to the question set. First, we created embeddings for each question using the question title and description. We used these embeddings to cluster related questions, and used an LLM to check for true duplicates within each cluster. For the embedding model, we used OpenAI's text-embedding-3-large, and clustered with DBSCAN with a cosine similiarity of 0.85. For the final LLM check, we used \claudehaiku{}. The prompt for the LLM check can be found in the Appendix \ref{sec:appendix-deduplication-prompt}.

\section{Question Resolver}
\label{sec:question-resolution}

We resolve questions using an ensemble of LLM-powered agents with internet access. Specifically, we employ three copies of a \geminithreepro{}-powered agent (one with a different prompt from the other two). For questions where these three agents do not reach unanimous agreement on the resolution, we run an additional \claudeopus{}-powered agent as a tiebreaker. See Section~\ref{sec:appendix-question-resolution-prompt} for the prompt used to resolve questions.

\section{Evaluation Logistics}
\label{sec:evaluation-methodology}

In order to evaluate our system, we generated, forecasted on, and resolved a set of 1499 questions. Question generation took place between September 27 and October 2, 2025. We prompted our system to generate questions resolving between October 15 and December 31, 2025, by specifying this requirement in both the proto-question generation and question refinement prompts (see Appendices~\ref{sec:appendix-prompt-seed-to-proto-question} and~\ref{sec:appendix-prompt-proto-question-to-refined-question}).

We obtained seeds from three sources, which we deemed likely to lead to valuable and consequential questions. The first source was a set of rationales of revenue forecasts for S\&P 500 companies \ifOP{made for \url{https://platform.stockfisher.app/}, a stock picking tool that}\ifICLR{from a stock forecasting platform that} uses long-term forecasts of company fundamentals to identify stocks for long-term investing. The second was GDELT (\url{https://www.gdeltproject.org/}), a project that collects and analyzes news and events from around the world. The third was Media Cloud (\url{https://www.mediacloud.org/}), another project that collects global news articles. The GDELT and Media Cloud APIs return a set of URLs which we then fetched the content of in a second step.
In total, we generated 500 seeds from the \ifOP{Stockfisher}\ifICLR{stock forecasting platform's} revenue forecast rationales (one for every S\&P 500 company), 200 seeds from Media Cloud URLs, and 1800 seeds from GDELT URLs.

For GDELT, we queried for article URLs referencing events from the two weeks preceding seed generation. The query sampled events with diversity across event types and geographies without filtering by specific topics. For Media Cloud, we used the Media Cloud API to retrieve articles from the preceding five days, again without topic-specific filtering. The full content of each retrieved article was used as a seed. For \ifOP{Stockfisher}\ifICLR{the stock forecasting platform}, we used the 2030 and 2035 revenue forecast rationales generated for each S\&P 500 company.

From these 2500 seeds, the pipeline generated 14073 proto-questions, which were then refined into 14073 forecasting questions. After filtering and deduplication, 1499 questions remained for evaluation.

Forecasting was performed in two stages. In stage 1, we gathered evidence from the live web for each question using a ReAct agent equipped with web search tools. In stage 2, we assigned probabilities to question outcomes using a single LLM call, with the web research in context. This separation allowed us to run models released after the initial research was conducted and to experiment with different forecasting prompts, without introducing bias from additional information becoming available or losing questions as they resolve over time. We ultimately made use of this possibility: since \geminithreepro{} was released close to the end of 2025, we based \geminithreepro{} forecasting on \gptfive{} research.

The research stage took place between October 21 and October 27, 2025 (main research on October 21--23, with retries for failed queries on October 27). Forecasting took place on different dates for different models. We included a fake date---the date that the research was carried out---in models' system prompts during forecasting.
Question resolution took place between January 15 and January 22, 2026.

\section{Results}
\label{sec:results}

\subsection{Overview}

We evaluate our system by checking the generated questions against the five desiderata outlined in the introduction via several complementary approaches, where each approach touches on one or more of the criteria. First, we cluster questions and assign topic labels with LLMs to explore subject-matter composition and intra-cluster diversity. Second, expert forecasters rate a subset of questions for resolvability and non-triviality. Third, we manually resolve a subset of questions, compare the annulment rate against human-written Metaculus questions, and measure the accuracy of our automatic resolver against human judgments. Fourth, to test difficulty, we resolve our complete question set automatically and forecast on it with various models, verifying that more capable models achieve better Brier scores, and that additional research effort via subquestion decomposition yields performance gains---demonstrating that the benchmark rewards intelligence and effort at high capability levels. Finally, we document qualitative patterns and limitations.
\begin{table}[tbp]
  \centering
  \caption{Distribution of generated questions by topic.}
  \begin{tabular}{lrr}
    \toprule
    \textbf{Topic} & \textbf{Count} & \textbf{\%} \\
    \midrule
    Regulatory and policy actions & 337 & 22.5\% \\
    US government and public policy & 189 & 12.6\% \\
    Macroeconomics and markets & 176 & 11.7\% \\
    International security and diplomacy & 168 & 11.2\% \\
    Court cases and investigations & 150 & 10.0\% \\
    Gaza war diplomacy & 121 & 8.1\% \\
    Sports and politics forecasts & 97 & 6.5\% \\
    Atlantic hurricanes and extreme weather & 84 & 5.6\% \\
    Iran nuclear program oversight & 61 & 4.1\% \\
    Commercial space launches and regulation & 58 & 3.9\% \\
    COP30 climate negotiations & 29 & 1.9\% \\
    Elections and vote outcomes & 29 & 1.9\% \\
    \midrule
    \textbf{Total} & \textbf{1499} & \textbf{100.0\%} \\
    \bottomrule
  \end{tabular}
  \label{tab:topic-distribution}
\end{table}

\subsection{Question Set Composition}

To get a sense for the composition of the generated question set and assess consequentialness, we clustered questions by topic using embeddings from text-embedding-3-large and $k$-means clustering, then labeled each cluster using \gptfivetwo{}. Table~\ref{tab:topic-distribution} shows the distribution across 12 identified topics; example questions from each cluster are provided in Appendix~\ref{sec:appendix-example-questions}.

The cluster labels confirm that our questions address substantive domains where forecasting capability matters: geopolitics (Gaza, Iran, international diplomacy), policy and regulation (US government policy, regulatory actions), economics (macroeconomics and markets), law (court cases and investigations), and infrastructure (space launches, extreme weather). While our news-seeded approach naturally emphasizes current affairs over science or technology, these topics represent areas where accurate forecasting has clear real-world value.

To assess diversity, we examined whether questions within each cluster are genuinely distinct or merely trivial variations of one another. We sampled 15 random pairs of questions within each cluster and used \gptfivetwo{} to rate their similarity on a 1--4 scale (1 = completely different entities/events; 2 = some overlap but asking different things; 3 = same entities/events with minor variations such as dates or thresholds; 4 = duplicates). The overall mean similarity was 1.32: 70\% of pairs were rated completely different, 28\% somewhat different, and only 2\% quite similar. No duplicates were found, validating our deduplication process (which used a different model and clustering algorithm). Even the most homogeneous clusters (COP30 climate negotiations, Iran nuclear oversight) averaged below 2.0, indicating that questions within topics address genuinely distinct events rather than trivial variations. Full per-topic results and the scoring prompt are provided in Appendix~\ref{sec:appendix-intra-cluster-diversity}.

\subsection{Expert Forecaster Ratings}

We contracted external expert forecasters to give their subjective evaluations of a subset of the generated questions, primarily to assess resolvability, and also touching on difficulty. Forecasters were asked to assign each question to one of three categories:

\begin{itemize}
  \item "accept"
  \item "soft reject": The question can technically resolve, but it is trivial or otherwise uninteresting.
  \item "hard reject": The question is ambiguous or otherwise flawed, and is unlikely to be resolved successfully.
\end{itemize}

On a set of 149 questions, 122 (75.2\%) questions were rated "accept", 25 (16.8\%) questions were rated "soft reject", and 12 (8.1\%) questions were rated "hard reject".

\subsection{Resolution Statistics}
\label{sec:results-resolution-statistics}

We resolved the 1499 generated questions using the ensemble approach described in Section~\ref{sec:question-resolution}.

\paragraph{Automated results.}
Of the 1499 questions, 436 (29.1\%) resolved positively, 1058 (70.6\%) resolved negatively, and 5 (0.3\%) were annulled by the question resolver. The three \geminithreepro{} agents did not reach unanimous agreement on 131 questions; the disagreements among them were fairly correlated, with 70 pairwise disagreements between the two identically-prompted copies versus 100--102 disagreements involving the differently-prompted agent. The \claudeopus{} tiebreaker agreed with the Gemini majority 74\% of the time when one existed. Notably, the \geminithreepro{} agents alone would have annulled zero questions; all 5 annulments came from \claudeopus{} on disputed questions. This reflects a bias in \geminithreepro{} away from annulment---it tends to accept Google search snippets at face value without verifying whether sources are admissible under the resolution criteria.

\paragraph{Human verification.}
To assess accuracy, we contracted an expert forecaster to carefully resolve a uniformly random subset of 100 questions. Comparing automated resolutions to this ground truth, we found 4 errors: 2 wrong binary resolutions (1 \yes{} that should have been \no{}, 1 \no{} that should have been \yes{}) and 2 missed annulments (questions resolved \yes{} or \no{} that should have been annulled). In total, the human evaluator determined that 3 of the 100 questions should be annulled, of which the automated system correctly identified 1. All three cases involved data sources that were unavailable: one due to the US government shutdown halting data collection (an external circumstance rather than a question quality issue), and two due to flaws in question generation where the admissible resolution sources did not publish the required data.

Manual resolution---even with AI assistance but with careful verification of outputs---required an average of 13 minutes per question (see Figure~\ref{fig:resolution-time} in the Appendix).

\paragraph{Inferred rates.}
Based on the human-verified sample and a uniform Beta(1,1) prior, we estimate:
\begin{itemize}[nosep]
  \item \textbf{Question quality:} Approximately 3.9\% of questions generated by our pipeline require annulment (95\% CI: [1.1\%, 8.4\%]) (based on 3 out of 100 annulled questions in the manually verified sample).
  \item \textbf{Resolution accuracy:} The automated resolver has a 4.9\% error rate (95\% CI: [1.6\%, 9.8\%]) (based on 4 wrong resolutions out of 100 in the manually verified sample).
\end{itemize}

To contextualize these rates, we compared against Metaculus, a leading human-curated forecasting platform. We find a striking similarity to Metaculus' base rate of approximately 30\% positive resolutions for binary questions. More notably, our estimated annulment rate of $\sim$4\% compares favorably to Metaculus' historical rate of approximately 8\%. That our automated system matches or exceeds human-curated question quality represents a strong validation of the approach.

\subsection{Forecasting-based Evaluation}
\label{sec:results-forecasting-based-evaluation}

If our generated questions sets are to be useful benchmarks of forecasting ability, it's important that greater intelligence yields better performance on generated question sets. We probe the extent to which our evaluation set exhibits this property by comparing models of varying capability (see Appendix~\ref{sec:appendix-model-details} for model specifications). Specifically, we expect larger models to outperform their smaller within-family counterparts (e.g., \gptfive{} vs \gptfivemini{}), and newer, more capable frontier models to outperform older ones.

We evaluated forecasting performance across five model combinations, separating the research phase (information gathering) from the forecasting phase (probability estimation).

The research stage was performed using a ReAct agent driven by the model under evaluation. The agent researched the live web via a Google Search tool, together with a tool for answering questions about documents found at a given URL (see prompt in Appendix~\ref{sec:appendix-prompt-final-forecast-research}). Forecasting was performed via a single LLM call with the research results in context (see prompt in Appendix~\ref{sec:appendix-prompt-final-forecast-probability}). Table~\ref{tab:model-scores} presents Brier scores with 95\% bootstrap confidence intervals (10,000 iterations). 

\begin{table}[tbp]
  \centering
  \caption{Forecasting performance by model combination. Forecast indicates the model that produced the final probability; Research indicates the model used for information gathering. Lower Brier scores indicate better performance; 95\% CIs computed via bootstrap (10,000 iterations). Calibration measures deviation from perfect calibration (lower is better); Refinement measures discriminative ability (higher is better). The bottom two rows show the subquestion decomposition experiment on a subset of questions.}
  \small
  \setlength{\tabcolsep}{4pt}
  \begin{tabular}{@{}lllcccc@{}}
    \toprule
    \textbf{Forecast} & \textbf{Research} & \textbf{Set} & \textbf{Brier} & \textbf{95\% CI} & \textbf{Calibration} & \textbf{Refinement} \\
    \midrule
    \geminithreepro{} & \gptfive{} & full & \textbf{0.134} & [0.123, 0.146] & \textbf{0.013} & \textbf{0.085} \\
    \gptfive{} & \gptfive{} & full & 0.149 & [0.141, 0.157] & 0.018 & 0.076 \\
    \gptfivemini{} & \gptfive{} & full & 0.155 & [0.146, 0.163] & 0.015 & 0.067 \\
    \geminitwofivepro{} & \geminitwofivepro{} & full & 0.165 & [0.155, 0.176] & 0.022 & 0.063 \\
    \geminitwofiveflash{} & \geminitwofivepro{} & full & 0.179 & [0.168, 0.191] & 0.026 & 0.054 \\
    \midrule
    \geminithreepro{} & \gptfive{} + subQ & subset & \textbf{0.132} & [0.115, 0.149] & 0.008 & 0.083 \\
    \geminithreepro{} & \gptfive{} & subset & 0.141 & [0.122, 0.161] & 0.026 & 0.091 \\
    \bottomrule
  \end{tabular}
  \label{tab:model-scores}
\end{table}

The results confirm our expectations: larger models outperform their smaller within-family counterparts (\gptfive{} beats \gptfivemini{}; \geminitwofivepro{} beats \geminitwofiveflash{}), and the newest frontier model, \geminithreepro{}, achieves the best performance overall, showing that reward from our generated question set increases with intelligence.
The rankings are highly stable under bootstrap resampling (10,000 iterations): \geminithreepro{} ranks first in 100\% of samples, \gptfive{} ranks second in 99.1\%, and the ordering within each model family is consistent. \gptfive{} beats \gptfivemini{} in 99.1\% of bootstrap samples, and \geminitwofivepro{} beats \geminitwofiveflash{} in 100\%.

To explore forecasting performance more deeply, we decompose the Brier score into calibration and refinement components. By binning forecasts into $K$ bins ($K = 10$ in our analysis), we obtain:
\begin{equation}
  \text{Brier} = \underbrace{\frac{1}{N} \sum_{k=1}^{K} n_k (f_k - o_k)^2}_{\text{Calibration}} - \underbrace{\frac{1}{N} \sum_{k=1}^{K} n_k (o_k - \bar{o})^2}_{\text{Refinement}} + \underbrace{\bar{o}(1 - \bar{o})}_{\text{Uncertainty}}
\end{equation}
where $N$ is the total number of forecasts, $n_k$ is the count in bin $k$, $f_k$ is the mean forecast in bin $k$, $o_k$ is the observed frequency of positive outcomes in bin $k$, and $\bar{o}$ is the overall base rate. Calibration measures how well forecast probabilities match observed frequencies (lower is better). Refinement measures the forecaster's ability to discriminate between outcomes (higher is better). Uncertainty depends only on the base rate and is constant across models.

The decomposition reveals that the expected ranking generally holds across both components: better-performing models tend to Pareto dominate weaker ones, achieving both lower calibration error and higher refinement. The exception is \gptfive{} versus \gptfivemini{}, where \gptfivemini{} is slightly better calibrated (0.015 vs 0.018) but has notably worse refinement (0.067 vs 0.076).

Reliability diagrams visualizing calibration for each model are provided in Appendix~\ref{sec:appendix-forecasting-figures} (Figure~\ref{fig:reliability-diagram}). As an additional sanity check on question quality, Figure~\ref{fig:forecast-distributions} in the same appendix shows the distribution of forecasts produced by each model across the benchmark. The questions elicit forecasts spanning the full probability range, with models producing mean forecasts between 0.37 and 0.42. Given the base rate of 29.2\%, all models exhibit some bias toward predicting \yes{} outcomes, though the distributions remain well-spread rather than concentrated at extreme values.

\paragraph{Subquestion decomposition.}
To probe our benchmark's ability to reward increased intelligence and effort beyond that of a ReAct forecaster driven by a leading frontier model, we conducted an experiment where higher-effort research was used as the basis for forecasts. Specifically, we randomly sampled 500 questions from our set and, for each, used a ReAct-style agent to generate 3--5 subquestions designed to be informative about the top-level question (see prompt in Appendix~\ref{sec:appendix-prompt-subquestion-decomposition}). We then researched and forecasted on these subquestions using \gptfive{}, and added the subforecasts and their accompanying research to the research context, before making the final forecast using \geminithreepro{} (see prompt in Appendix~\ref{sec:appendix-prompt-subquestion-forecast}). Research and forecasting for the subquestions was conducted two days after we captured research for the top-level questions (October 23 vs October 21, 2025, with some retries on October 27.).

Comparing \geminithreepro{} forecasts with and without the subquestion decomposition on the sampled questions, we find that subquestion decomposition improves forecasting performance: Brier score decreases from 0.141 (95\% bootstrap CI: [0.122, 0.161]) to 0.132 (95\% bootstrap CI: [0.115, 0.149]). In bootstrap resampling, the subquestion-augmented forecasts outperform the baseline in 94.4\% of samples (10,000 iterations).

\subsection{Qualitative Observations}

Beyond the quantitative metrics above, manually inspecting questions and resolutions revealed several patterns worth noting. We discuss characteristics of the generated questions that affect their difficulty and resolvability, systematic limitations of LLM-based resolution, and the impact of model choice on resolution quality.

\subsubsection{Observations on question characteristics}


A notable pattern emerged around questions of the form ``\textit{[Big institution announced intentions to do X] $\rightarrow$ will X happen within the next 3 months?}'' These questions almost invariably resolve \no{} because large institutions, particularly governmental bodies like the EU, operate on extended timelines. Examples include questions about Canada Post and CUPW ratifying collective agreements by specific dates, or various EU policy implementation deadlines.

Many questions require establishing that something \textit{did not} happen for a successful \no{} resolution. For sufficiently high-profile events, absence of evidence can reasonably serve as evidence of absence---we would certainly hear about a US attack on Greenland, for instance---and they are easily resolved. However, this heuristic fails for more obscure questions about technical or bureaucratic matters, complicating the resolution process for weaker/lazier models: For example, questions like ``\textit{Will OFAC add at least five Venezuela-related persons or entities to the SDN List under E.O. 13224 and/or E.O. 14059 between 2025-10-15 and 2025-12-31 (UTC)?}'' cannot be reliably resolved through this approach without substantial effort, and is indeed not resolved correctly by the \geminithreepro{} agents, as the absence of Google results suggesting that this occurred does not imply that it did not occur.

\subsubsection{Observations on LLM-based resolution}

Using plurality voting among multiple agents can mitigate issues with question resolutions, as described in Section~\ref{sec:results-resolution-statistics}. The three Gemini 3 Pro agents disagreed on 131 of 1499 questions (8.7\%), and the Opus 4.5 tiebreaker overrode the Gemini majority in only 26\% of those cases. However, this approach cannot entirely eliminate correlated failures across models.

We identified several systematic limitations affecting most current LLM agents (including Claude Code, ChatGPT, and various deep research tools---essentially all tools we had some experience with except OpenAI's Operator):
\begin{itemize}[nosep]
  \item Finding information that requires filling out forms, search bars, or other interactive inputs
  \item Accessing data that requires clicking to dynamically load content
  \item Reliably extracting specific values from long PDF documents
\end{itemize}

Despite these limitations, agents generally perform well because most data of interest is referenced in multiple locations. However, they occasionally misinterpret information from secondary sources.

The most challenging questions fall into two categories: (1) those requiring interaction with databases, such as the OFAC SDN List example above, and (2) those requiring verification that something \textit{did not} happen when that event would not have been widely reported. An example of the latter: ``\textit{Will the EU/Italy formalize financing or publish a tender for the Blue-Raman East Africa extension by 23:59 UTC on December 31, 2025?}''

\subsubsection{Model choice matters substantially for question resolution}

The choice of model significantly impacts resolution quality. For the question ``\textit{Between 2025-10-15 and 2025-12-31 (UTC), will USDOT list any FY2025 Safe Streets and Roads for All (SS4A) grant awards on its SS4A awards pages?}'', \claudeopus{} either recognizes that verification is technically impossible (depending on prompt) or manages to use the Wayback Machine to obtain snapshots from the requested period for rigorous resolution. \geminithreepro{}, by contrast, simply checks the current website, notes a ``last updated'' timestamp within the relevant window, and resolves without further verification.

\geminithreepro{} also exhibits poor instruction-following, requiring practitioners to effectively work within its default behaviors rather than customizing through prompting.

\section{Discussion}
\label{sec:discussion}

We have presented a system for automatically generating forecasting questions that are unambiguously resolvable, difficult, consequential, and diverse. Our evaluation demonstrates that the generated questions reward increased intelligence and effort: larger models consistently outperform their smaller within-family counterparts, and the newest frontier models achieve the best performance. The questions resolve at rates comparable to or better than those observed on leading forecasting platforms, with an estimated annulment rate of 3.9\% (95\% CI: [1.1\%, 8.4\%])---comparable to Metaculus' historical rate of $\sim$8\%. Topic clustering confirms that the questions span a broad range of consequential domains, from international security and monetary policy to climate diplomacy and legal developments, while intra-cluster analysis shows that questions within each topic remain diverse rather than repetitive. Together, these results suggest that automated question generation can produce evaluation sets of sufficient quality for accurately benchmarking cutting-edge forecasting systems.

Several directions for future work merit investigation. A deeper analysis of question interestingness and difficulty would help refine the generation process---understanding what makes certain questions more informative or challenging could guide the system toward higher-value outputs. Focusing generation on particularly high-impact domains, such as biosecurity, AI development, or geopolitical stability, would increase the practical relevance of generated benchmarks. Perhaps most promising is the extension to conditional questions (e.g., ``If policy X is enacted, will outcome Y occur?''), which would add complexity and be especially valuable for decision-making applications where understanding the consequences of potential interventions is paramount.

\newpage

\ifOP{
  
\section*{Acknowledgments} 
This work was supported by Coefficient Giving.
}

\bibliographystyle{plainnat}
\bibliography{bib/references,bib/zotero}

\clearpage

\appendix
\section{Appendix}
\label{sec:appendix}

\subsection{Manual Resolution Time}
\label{sec:appendix-resolution-time}

\begin{figure}[htbp]
  \centering
  \includegraphics[width=0.7\textwidth]{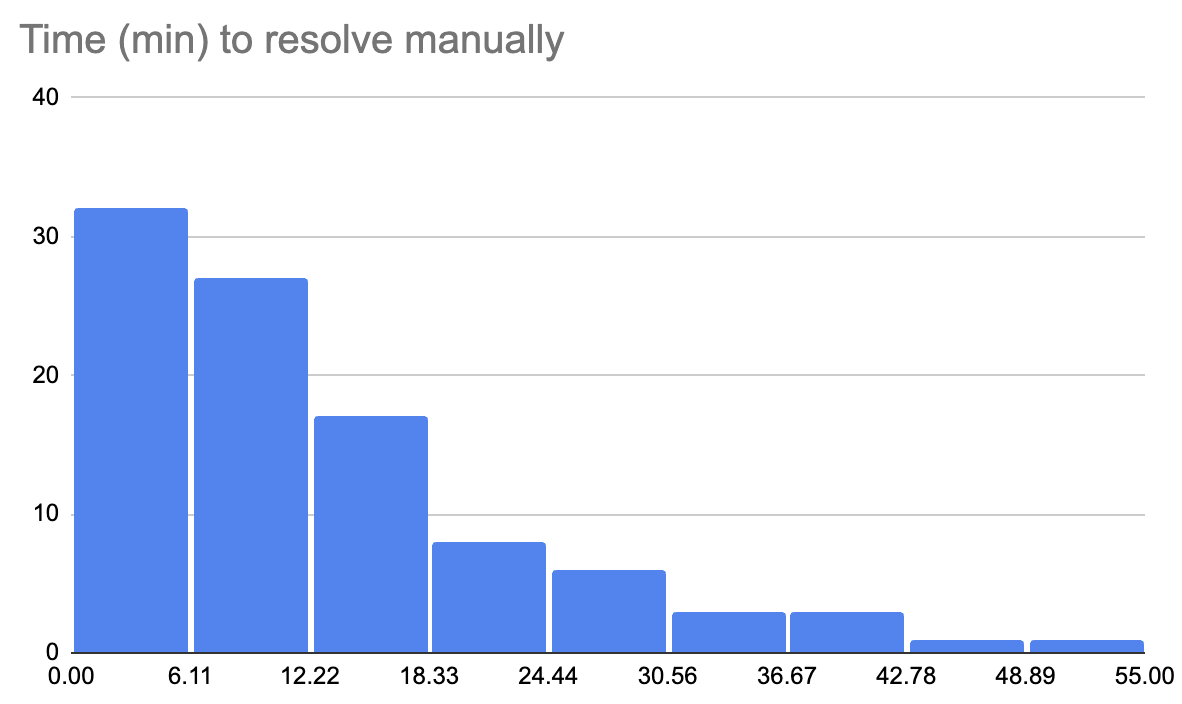}
  \caption{Distribution of time required to manually resolve questions with AI assistance. The median resolution time was approximately 6--12 minutes, with a long tail of more complex questions requiring up to 55 minutes.}
  \label{fig:resolution-time}
\end{figure}
\FloatBarrier

\subsection{Desirable Qualities of Forecasting Questions}
\label{sec:appendix-desirable-qualities}

Our goal is to create a set of questions that is as useful as possible for evaluating and discerning differences in the performance of AI systems on forecasting tasks. This implies a number of desirable qualities for both individual forecasting questions and for question sets as a whole. The following provides a more detailed operationalization of those qualities. We distinguish between ``primary qualities'' that we explicitly aimed to optimize for and ``secondary qualities'' that may be desirable but were not our primary focus.

\begin{table}[htbp]
  \centering
  \caption{Desirable qualities for forecasting questions and question sets.}
  \begingroup
  \renewcommand{\arraystretch}{1.3}
  \begin{tabularx}{\textwidth}{@{}>{\hsize=.12\hsize}X>{\hsize=.28\hsize}X>{\hsize=.6\hsize\arraybackslash}X@{}}
    \toprule
    \multicolumn{3}{c}{\textbf{Individual Question Qualities}} \\
    \cmidrule(r){1-3}
    \textbf{Type} & \textbf{Quality} & \textbf{Explanation} \\
    Primary & Unambiguously resolvable & Reduces noise and avoids wasteful effort. Enables automated evaluation of question outcomes. \\
    Primary & Discriminates forecasting skill & Better research and judgment should \textit{consistently} lead to better performance. Questions like ``forecast the roll of a fair die'' or ``Will Apple's stock price exceed \$X on date Y?'' violate this. Within reasonable limits, no forecaster should be expected to systematically outperform efficient market prices. \\
    Primary & Timely resolution & For practical evaluation purposes, questions should resolve within a reasonable timeframe. \\
    Primary & Non-extreme base rate & Forecasts should not trivially cluster around 0\% or 100\%, enabling meaningful calibration assessment and providing high information value. \\
    Secondary & Interesting & Interestingness is subjective and depends on the forecast consumer. Many highly relevant forecasting tasks involve repetitive research that may not be ``interesting'' in a colloquial sense. \\
    Secondary & Decision-relevant & Usefulness depends on the specific decision-maker and context. \\
    \addlinespace[8pt]
    \multicolumn{3}{c}{\textbf{Question Set Qualities}} \\
    \cmidrule(r){1-3}
    \textbf{Type} & \textbf{Quality} & \textbf{Explanation} \\
    Primary & Representative & Performance on the question set should be a valid proxy for performance on real-world forecasting tasks, at least in terms of the skills required. \\
    Primary & Diverse & Covers varied topics and question types to avoid overfitting to narrow domains. \\
    Primary & Appropriate difficulty & Questions should be (a) not so easy that all forecasters score similarly, (b) not so hard that all fail equally, and (c) challenging enough that top forecasters do not hit a ceiling. \\
    Primary & Sufficient size & Large enough to ensure adequate statistical power for detecting performance differences. \\
    Primary & Independent outcomes & Questions should be as uncorrelated as possible to increase statistical power and prevent forecasters from benefiting disproportionately from luck on correlated outcomes. \\
    Secondary & Controlled difficulty distribution & Having forecasts across different probability ranges aids calibration evaluation. \\
    Secondary & Varied time horizons & Different resolution dates enable evaluation across short- and long-term forecasting. \\
    \bottomrule
  \end{tabularx}
  \endgroup
  \label{tab:qualities}
\end{table}
\FloatBarrier

\paragraph{On question difficulty.} Forecasting difficulty can be decomposed into two components: \textit{research difficulty} (the ability to obtain better information) and \textit{judgment difficulty} (forming superior predictions given the same information). These dimensions are partially interchangeable---thorough research can reduce the need for judgment (e.g., looking up earthquake base rates rather than relying on intuition), while strong reasoning can sometimes substitute for research (e.g., deducing likely actions from first principles rather than finding explicit statements). Our question generation pipeline aims to produce questions that reward both capabilities.

\subsection{Example Questions by Topic}
\label{sec:appendix-example-questions}

The following examples illustrate the diversity of questions in each topic cluster identified in Section~\ref{tab:topic-distribution}.

\paragraph{Regulatory and policy actions} (337 questions)
\begin{itemize}[nosep]
  \item Will the U.S. FDA approve topical ocular reproxalap (Aldeyra Therapeutics) for the treatment of dry eye disease on or before December 31, 2025?
  \item Will the EU publish in the Official Journal (L series), by 31 December 2025, a Regulation on end-of-life vehicles that imposes a binding minimum recycled-plastic content requirement for new vehicle types?
  \item Between October 15 and December 31, 2025 (UTC), will the IMF Executive Board complete the second review of Pakistan's Extended Fund Facility (EFF) and approve a disbursement under the EFF?
  \item Between October 15 and December 31, 2025 (UTC), will the IMF Executive Board complete the Fifth Review under Sri Lanka's Extended Fund Facility (EFF)?
  \item UK: Will the Sentencing Bill (introduced 2 Sep 2025) receive Royal Assent on or before 31 Dec 2025?
\end{itemize}

\paragraph{US government and public policy} (189 questions)
\begin{itemize}[nosep]
  \item Will CDC-reported U.S. measles cases in 2025 reach at least 2,000 by December 31, 2025 (UTC)?
  \item If a federal government shutdown occurs between October 15 and December 31, 2025 (UTC), will at least three U.S. states fund operations to keep any National Park System units within their borders open with essential services during the shutdown?
  \item Will U.S. Customs and Border Protection (CBP) report at least 800 pounds of fentanyl seized in November 2025?
  \item Will Guatemala's INACIF-reported homicide count for January--November 2025 exceed the count for January--November 2024?
  \item Between October 15 and December 31, 2025 (UTC), will the U.S. federal government experience a shutdown of non-excepted operations lasting at least five consecutive calendar days?
\end{itemize}

\paragraph{Macroeconomics and markets} (176 questions)
\begin{itemize}[nosep]
  \item Will the December 2025 WASDE project U.S. 2025/26 corn ending stocks at $\leq$ 2.05 billion bushels?
  \item Italy BEV share: Will battery-electric vehicles be at least 6.5\% of new passenger-car registrations in Italy in November 2025?
  \item At the daily close on December 31, 2025, will Spain's 10-year government bond yield be lower than France's?
  \item In EIA's Electric Power Monthly, will coal's share of U.S. net electricity generation for October 2025 be at least 16.00\%?
  \item Will S\&P Dow Jones Indices announce the addition of StandardAero (NYSE: SARO) to the S\&P MidCap 400 by December 31, 2025?
\end{itemize}

\paragraph{International security and diplomacy} (168 questions)
\begin{itemize}[nosep]
  \item Will a NATO member state's military shoot down at least one Russian-state-origin drone over NATO territory between 2025-10-15 and 2025-12-31 (UTC)?
  \item Will Bulgaria have at least eight F-16 Block 70 aircraft physically arrive in-country by 2025-12-31 23:59:59 UTC?
  \item Will OFAC add at least five Venezuela-related persons or entities to the SDN List under E.O. 13224 and/or E.O. 14059 between 2025-10-15 and 2025-12-31 (UTC)?
  \item Will a NATO member's armed forces shoot down a crewed Russian military aircraft inside that same member's national airspace, in response to an unauthorized incursion, between 2025-10-15 and 2025-12-31 (UTC)?
  \item Will Poland or Romania officially confirm a Russian drone or missile incursion into their national airspace between 2025-10-15 and 2025-12-31 (UTC)?
\end{itemize}

\paragraph{Court cases and investigations} (150 questions)
\begin{itemize}[nosep]
  \item Will SCOTUS consolidate the two government birthright-citizenship cert petitions for merits consideration by December 31, 2025?
  \item Will the House or Senate Judiciary Committee formally announce an investigation into the late-September 2025 FBI agent firings between October 15 and December 31, 2025 (UTC)?
  \item Will DOJ publish both the Attorney General's written guidance and letter described in Section 3 of the Sept 25, 2025 White House executive order ``Saving TikTok While Protecting National Security'' by Dec 31, 2025?
  \item By December 31, 2025, will a superseding indictment be filed in United States v. Comey (E.D. Va., No. 1:25-cr-00272)?
  \item Moldova: Will a top leader of the Patriotic Electoral Bloc or Heart of Moldova be newly arrested or formally charged for illegal campaign financing between 2025-10-15 and 2025-12-31 (UTC)?
\end{itemize}

\paragraph{Gaza war diplomacy} (121 questions)
\begin{itemize}[nosep]
  \item Will New Zealand formally recognise the State of Palestine between 2025-10-01 and 2025-12-31 (UTC)?
  \item Between October 1 and December 31, 2025 (UTC), will Israel's Knesset be dissolved with an official election date set on or before May 31, 2026?
  \item Will at least 10 hostages held in the Gaza Strip be released alive as part of a publicly announced negotiated agreement between October 15 and December 31, 2025 (UTC)?
  \item Between Oct 15 and Dec 31, 2025 (UTC), will any national government other than Italy or Spain publicly announce or be credibly reported as deploying naval or coast guard assets to monitor, escort, or provide SAR support for a Gaza-bound civilian aid flotilla?
  \item In December 2025, will UN2720 ``Collected from crossings'' trucks total at least 2,170 ($\geq$70 per day on average)?
\end{itemize}

\paragraph{Sports and politics forecasts} (97 questions)
\begin{itemize}[nosep]
  \item On December 31, 2025 (after all scheduled NHL games that day), will the Toronto Maple Leafs lead the NHL Atlantic Division in points (applying the official NHL tie-breaking procedure)?
  \item Will Coco Gauff be named the WTA Player of the Year for 2025?
  \item LaLiga 2025/26: Will any Atl\'etico Madrid player be among the top three goal scorers (by goals) on LaLiga's official stats page at 23:59:59 CET on December 31, 2025?
  \item Will at least 19 APEC economy leaders attend the APEC Economic Leaders' Meeting in Gyeongju (Oct 31--Nov 1, 2025, KST)?
  \item Will Haas finish the 2025 Formula 1 season with at least 55 Constructors' Championship points?
\end{itemize}

\paragraph{Atlantic hurricanes and extreme weather} (84 questions)
\begin{itemize}[nosep]
  \item Will the National Hurricane Center's ``2025 North Atlantic Summary'' list at least 15 named storms as of 23:59 UTC on November 30, 2025?
  \item From 2025-10-15 through 2025-12-31 (UTC), will NOAA SWPC record at least 3 X-class solar flares?
  \item Will the 2025 North Atlantic hurricane season have at least 4 major hurricanes (Category 3+)?
  \item Will the NHC's final 2025 Atlantic ``Total ACE'' be at least 115 $\times$ 10\^{}4 kt\^{}2 by December 31, 2025?
  \item Will NOAA explicitly classify the 2025 Atlantic hurricane season as above-normal in an end-of-season communication published between Oct 15 and Dec 31, 2025?
\end{itemize}

\paragraph{Iran nuclear program oversight} (61 questions)
\begin{itemize}[nosep]
  \item Will Russia or China explicitly declare non-implementation of the reimposed UN Security Council sanctions on Iran between 2025-10-15 and 2025-12-31 (UTC)?
  \item Between October 15 and December 31, 2025 (UTC), will Iran formally notify withdrawal from the Nuclear Non-Proliferation Treaty (NPT) under Article X?
  \item Between Oct 15 and Dec 31, 2025 (UTC), will a recognized OSINT organization or major news outlet publish commercial satellite imagery analysis identifying a new tunnel entrance (adit) at the Kuh-e Kolang Gaz La (``Pickaxe Mountain'') underground complex near Natanz?
  \item Iran nuclear inspections: Will IAEA inspectors physically access Fordow (FFEP) or Natanz (FEP or PFEP) at least once between 2025-10-15 and 2025-12-31 (UTC)?
  \item By 23:59 UTC on 31 December 2025, will the IAEA explicitly state that continuity of knowledge over Iran's inventories of highly enriched uranium ($\geq$20\% U-235) has been restored?
\end{itemize}

\paragraph{Commercial space launches and regulation} (58 questions)
\begin{itemize}[nosep]
  \item Will Blue Origin's New Glenn lift off on a second orbital mission between 2025-10-15 and 2025-12-31 (UTC)?
  \item Will Blue Origin's New Glenn launch NASA's ESCAPADE mission between 00:00:00 UTC on October 15, 2025 and 23:59:59 UTC on December 31, 2025?
  \item By December 31, 2025 (23:59:59 UTC), will NASA publish on NASA.gov a schedule update stating that Artemis II will launch after April 2026 (i.e., a target date or ``no earlier than'' date of May 2026 or later)?
  \item Will the FAA publicly lift its 737 MAX production expansion freeze or explicitly raise the 38-per-month cap between October 15 and December 31, 2025?
  \item Will ISRO's uncrewed Gaganyaan-G1 mission lift off between 00:00 IST on 15 Oct 2025 and 23:59:59 IST on 31 Dec 2025?
\end{itemize}

\paragraph{COP30 climate negotiations} (29 questions)
\begin{itemize}[nosep]
  \item Will any COP30 or CMA.7 cover decision contain the exact phrase ``phase-out of fossil fuels''?
  \item Will CMA7 at COP30 adopt a decision establishing a final set of Global Goal on Adaptation (GGA) indicators, with the adopted decision published on the UNFCCC website by 23:59 UTC on December 31, 2025?
  \item Will cumulative pledges to the UNFCCC Fund for responding to Loss and Damage reach at least USD \$1.30 billion by 23:59 BRT on November 21, 2025?
  \item COP30 operations: Will the COP 30 closing plenary start on or after Sunday, 23 November 2025 (America/Belem local time)?
  \item By 31 December 2025, will at least 13 of the 19 G20 sovereign states plus the European Union have an NDC communication dated in 2025 recorded in the UNFCCC NDC Registry?
\end{itemize}

\paragraph{Elections and vote outcomes} (29 questions)
\begin{itemize}[nosep]
  \item Thiruvananthapuram Municipal Corporation 2025: Will the Bharatiya Janata Party (BJP) win at least 34 wards?
  \item Will Prashant Kishor's Jan Suraaj Party win at least one seat in the 2025 Bihar Legislative Assembly election?
  \item Will BJP, JD(U), LJP(RV), HAM(S), and RLM win at least 150 seats in the 2025 Bihar Assembly election?
  \item Will Nitish Kumar be the Chief Minister of Bihar on 31 December 2025?
  \item Will the LDF win more District Panchayat member seats than the UDF in Kerala's 2025 local body elections?
\end{itemize}

\subsection{Intra-Cluster Diversity Analysis}
\label{sec:appendix-intra-cluster-diversity}

To assess whether questions within each topic cluster are genuinely diverse or merely trivial variations, we sampled 15 random pairs of questions from each cluster and used \gptfivetwo{} to rate their similarity on a 1--4 scale:
\begin{itemize}[nosep]
  \item 1 = Completely different: no shared entities or events
  \item 2 = Somewhat different: some overlap in entities/events, but asking about different things
  \item 3 = Quite similar: same entities/events, differing only by minor variations (dates, thresholds)
  \item 4 = Duplicates: asking the same thing in different words
\end{itemize}

Table~\ref{tab:intra-cluster-diversity} shows the results by topic.

\begin{table}[htbp]
  \centering
  \caption{Intra-cluster diversity analysis. Lower mean similarity indicates greater diversity within the cluster.}
  \begin{tabular}{lrrrrr}
    \toprule
    \textbf{Topic} & \textbf{N} & \textbf{Mean} & \textbf{1s} & \textbf{2s} & \textbf{3s} \\
    \midrule
    Court cases and investigations & 150 & 1.00 & 15 & 0 & 0 \\
    Regulatory and policy actions & 337 & 1.00 & 15 & 0 & 0 \\
    Sports and politics forecasts & 97 & 1.00 & 15 & 0 & 0 \\
    US government and public policy & 189 & 1.07 & 14 & 1 & 0 \\
    Macroeconomics and markets & 176 & 1.13 & 13 & 2 & 0 \\
    International security and diplomacy & 168 & 1.20 & 12 & 3 & 0 \\
    Commercial space launches and regulation & 58 & 1.27 & 11 & 4 & 0 \\
    Elections and vote outcomes & 29 & 1.33 & 10 & 5 & 0 \\
    Gaza war diplomacy & 121 & 1.53 & 7 & 8 & 0 \\
    Atlantic hurricanes and extreme weather & 84 & 1.60 & 7 & 7 & 1 \\
    Iran nuclear program oversight & 61 & 1.80 & 4 & 10 & 1 \\
    COP30 climate negotiations & 29 & 1.93 & 3 & 10 & 2 \\
    \midrule
    \textbf{Overall} & \textbf{1499} & \textbf{1.32} & \textbf{126} & \textbf{50} & \textbf{4} \\
    \bottomrule
  \end{tabular}
  \label{tab:intra-cluster-diversity}
\end{table}
\FloatBarrier

No duplicate pairs (score 4) were found among the 180 evaluated pairs. The clusters with highest similarity---COP30 climate negotiations and Iran nuclear oversight---reflect inherently narrow domains where questions naturally share entities, yet even these average below 2.0 (``somewhat different'').

\paragraph{Similarity scoring prompt.} The following prompt was used for pairwise similarity assessment:

\begin{tcolorbox}[
    breakable,
    width=\textwidth,
    left=6pt,
    right=6pt,
    top=4pt,
    bottom=4pt,
    boxrule=0.5pt,
    colback=gray!10,
    colframe=gray!50,
    title=Intra-Cluster Similarity Scoring Prompt
]
\small
\texttt{Compare these two forecasting questions and rate their similarity on a scale of 1-4:}

\texttt{Question A: [question text]}

\texttt{Question B: [question text]}

\texttt{Similarity scale:}\\
\texttt{1 = Completely different: they never refer to the same entity or event}\\
\texttt{2 = Somewhat different: some overlap in entities or events referenced, but asking about different things}\\
\texttt{3 = Quite similar: the entities and events are essentially the same, differing only by minor variations (e.g., different dates, different numeric thresholds)}\\
\texttt{4 = Frank duplicates: asking the same thing in different words}

\texttt{Respond with ONLY a single digit (1, 2, 3, or 4).}
\end{tcolorbox}

\subsection{Model Details}
\label{sec:appendix-model-details}

Table~\ref{tab:model-details} provides the specifications of all language models used in our experiments.

\begin{table}[htbp]
  \centering
  \caption{Language model specifications. Reasoning/thinking settings control extended reasoning capabilities where available.}
  \begin{tabular}{llll}
    \toprule
    \textbf{Display Name} & \textbf{Model Identifier} & \textbf{Reasoning/Thinking} & \textbf{Temperature} \\
    \midrule
    \gptfive{} & gpt-5 & low & 1 \\
    \gptfivemini{} & gpt-5-mini & medium & 1 \\
    \gptfivetwo{} & gpt-5.2 & low & 1 \\
    \geminithreepro{} & gemini-3-pro-preview & low & 1 \\
    \geminitwofivepro{} & gemini-2.5-pro & enabled & default \\
    \geminitwofiveflash{} & gemini-2.5-flash & disabled & default \\
    \claudehaiku{} & claude-haiku-4.5 & default & 1 \\
    \claudeopus{} & claude-opus-4.5 & low & 1 \\
    \bottomrule
  \end{tabular}
  \label{tab:model-details}
\end{table}
\FloatBarrier

\subsection{Forecasting Evaluation Figures}
\label{sec:appendix-forecasting-figures}

\begin{figure}[htbp]
  \centering
  \includegraphics[width=\textwidth]{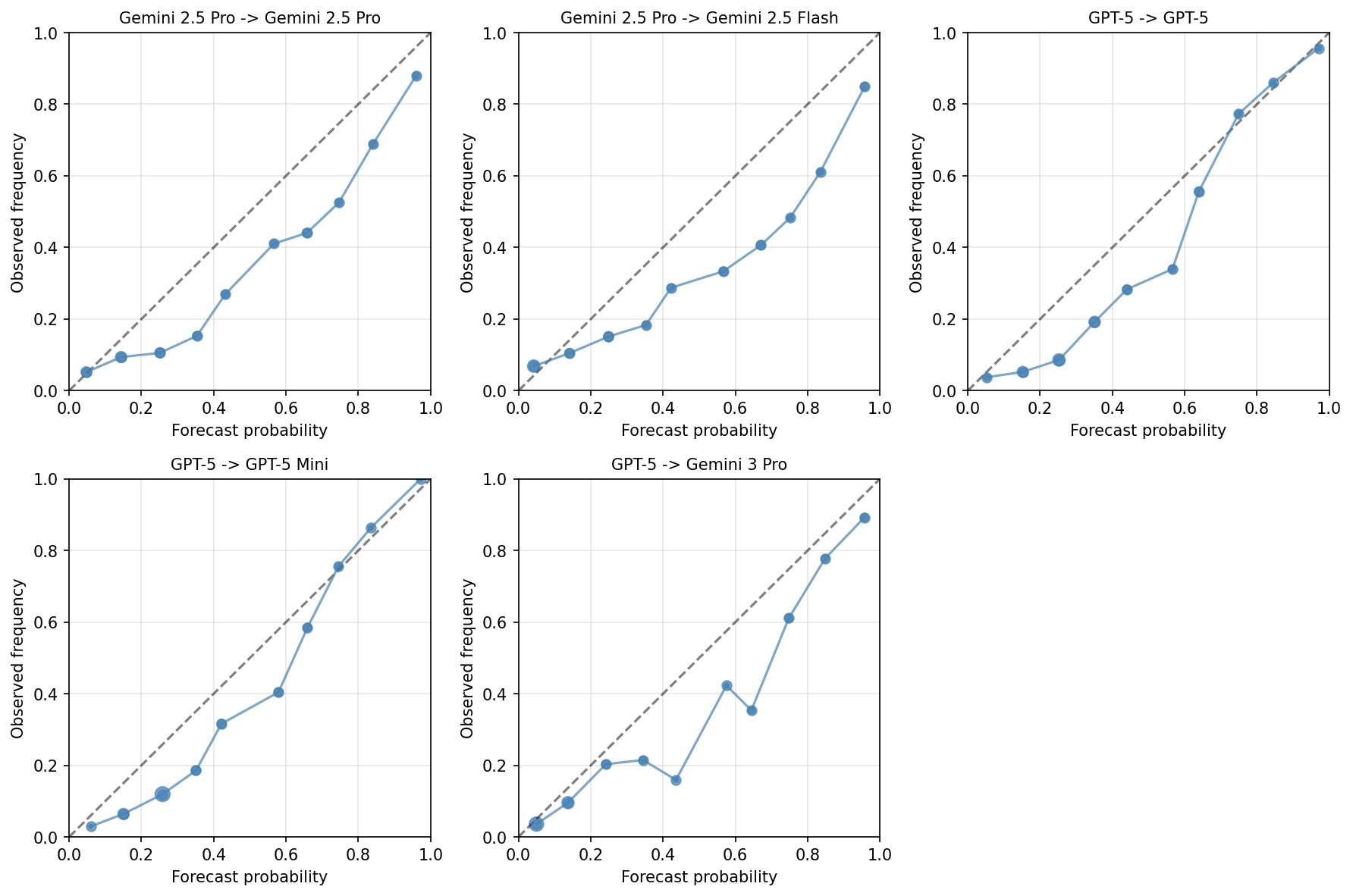}
  \caption{Reliability diagrams for each model combination. The diagonal line represents perfect calibration. Point size is proportional to the number of forecasts in each bin. Points above the diagonal indicate underconfidence (forecasts lower than observed frequencies), while points below indicate overconfidence.}
  \label{fig:reliability-diagram}
\end{figure}

\begin{figure}[htbp]
  \centering
  \includegraphics[width=\textwidth]{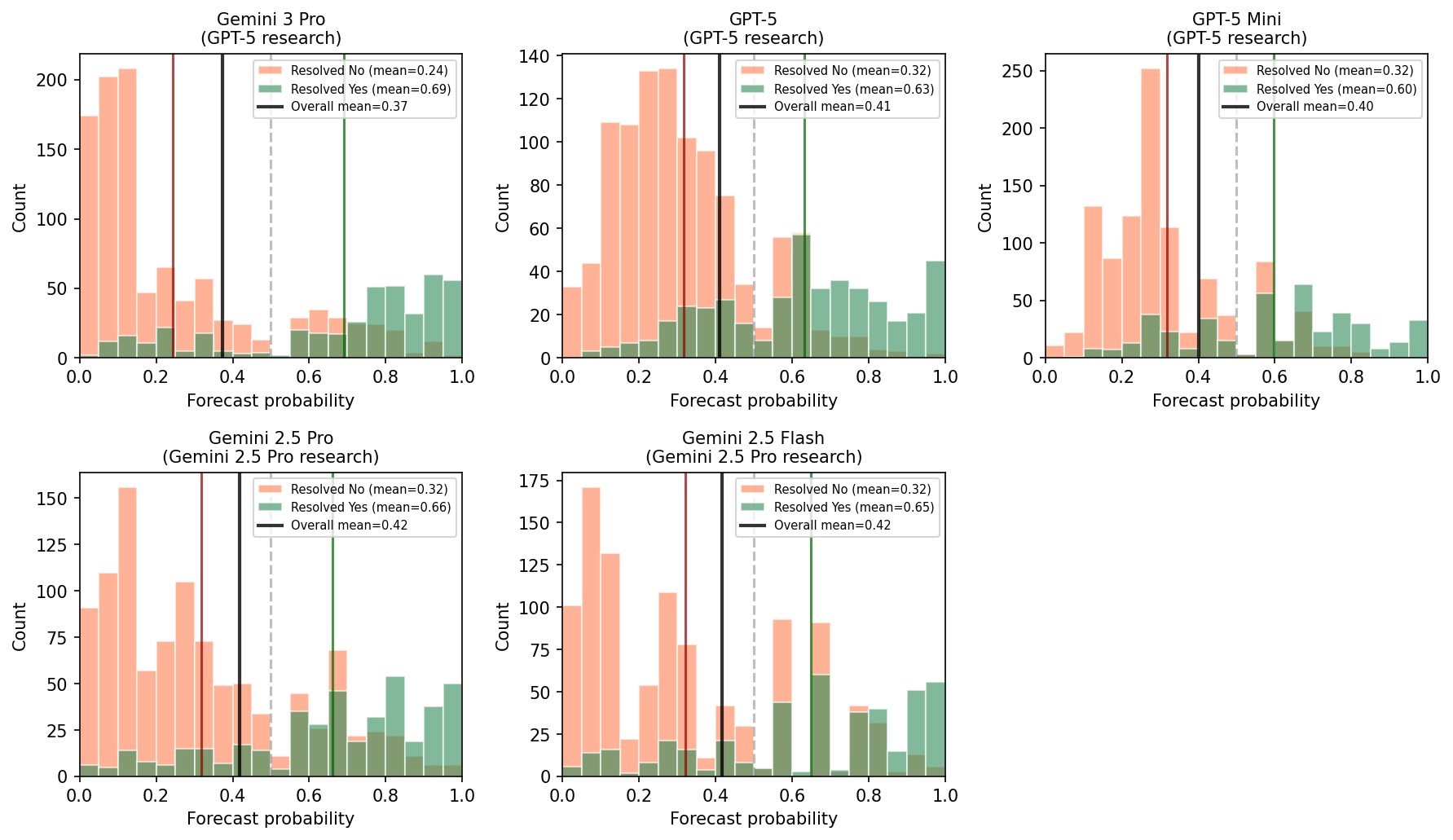}
  \caption{Distribution of forecasts by model combination, separated by resolution outcome. Green bars show forecasts for questions that resolved \yes{}; coral bars show forecasts for questions that resolved \no{} (annulled questions excluded). Vertical lines indicate mean forecasts: green for \yes{} outcomes, red for \no{} outcomes, and black for overall mean. Dashed gray lines mark 50\%. All models show good separation between forecasts for \yes{} and \no{} outcomes, indicating discriminative ability.}
  \label{fig:forecast-distributions}
\end{figure}
\FloatBarrier

\subsection{Question Generation Prompts}
\label{sec:appendix-question-generation-prompts}

\subsection{Agent Prompt to Generate Proto-Questions from a Question Seed}
\label{sec:appendix-prompt-seed-to-proto-question}

\begin{tcolorbox}[
    breakable,
    width=\textwidth,
    left=10pt,
    right=10pt,
    top=6pt,
    bottom=6pt,
    boxrule=0.5pt,
    colback=gray!10,
    colframe=gray!50,
    title=Agent Prompt to Generate Proto-Questions from a Question Seed
]
\begin{lstlisting}[
    breaklines=true,
    breakatwhitespace=true,
    basicstyle=\small\ttfamily,
    frame=single,
    framesep=2mm,
    xleftmargin=2mm,
    xrightmargin=2mm,
    backgroundcolor=\color{gray!10},
    showstringspaces=false,
    columns=flexible,
    literate={https://}{https://\allowbreak}1
            {http://}{http://\allowbreak}1
            {www.}{www.\allowbreak}1
            {.com}{.com\allowbreak}1
            {.org}{.org\allowbreak}1
            {.net}{.net\allowbreak}1
            {.edu}{.edu\allowbreak}1
            {.gov}{.gov\allowbreak}1
            {/}{/\allowbreak}1
            {?}{?\allowbreak}1
            {&}{&\allowbreak}1
            {=}{=\allowbreak}1
]
I am an admin at Metaculus, and I want to design a forecasting tournament. I collected a bunch of news articles and am now going through the list 1 by 1 in search of inspiration for some forecasting questions. The overall process for my forecasting tournament looks like this:

1. do some web research to come up with a set of "proto-questions",
2. turn each of these proto-questions into unambiguous and objectively resolvable, binary forecasting questions that resolve October 15, 2025 and the end of 2025. 
3. run a forecasting tournament on these questions to identify the best forecasters.

I want you to do step 1. Based on the given context/input, return 0-7 proto-questions (+ your own rationale for why you believe this is a good question). You can take the input as inspiration, but please do some research yourself to identify suitable questions. The question(s) should have the following properties:

- They ought to be questions about the near-term future and should resolve between October 15, and December 31, 2025. Pay attention to today's date and don't suggest questions that resolve before October 15 (or where the information is essentially known before October 15). 
- They need to be actually resolvable. This means: we have to be able to locate a source of truth that tells us whether the answer to the question is yes or no. This source has to actually exist on December 31, the date we want to resolve questions. This is important because sometimes conclusive data will only be available much later. We therefore need to make sure to only look at things that can be known on December 31, 2025.
- They ought to be maximally independent in the sense that a forecaster can't just identify a small number of confounders to effectively "bet on". Questions should be diverse and ask about different things
- They ought to have "high entropy", i.e., they should be non-trivial questions, with answers that are not almost certainly true or false.
- A good forecast for these questions should be between 5% and 95% probability. This means that you need to make sure that the question you propose can plausibly resolve both yes and no. Avoid things that will almost certainly go "yes" or "no" and also keep this in mind when setting any kinds of thresholds (these should be designed to keep things interesting)
- There could be two good forecasters whose forecasts differ by at least 20 percentage points; so no "clear" questions where every reasonable forecaster would forecast ~0% or ~100% and no overly base-rate-ish questions with a "correct" answer.
- Questions should not be questions where one can't do much better than e.g., looking up an existing forecast or, for example, the price of an Option on some exchange.

\end{lstlisting}
\end{tcolorbox}
\subsection{Agent Prompt to Refine a Question into a Fully Fledged Forecasting Question}
\label{sec:appendix-prompt-proto-question-to-refined-question}

\begin{tcolorbox}[
    breakable,
    width=\textwidth,
    left=10pt,
    right=10pt,
    top=6pt,
    bottom=6pt,
    boxrule=0.5pt,
    colback=gray!10,
    colframe=gray!50,
    title=Agent Prompt to Refine a Question into a Fully Fledged Forecasting Question
]
\begin{lstlisting}[
    breaklines=true,
    breakatwhitespace=true,
    basicstyle=\small\ttfamily,
    frame=single,
    framesep=2mm,
    xleftmargin=2mm,
    xrightmargin=2mm,
    backgroundcolor=\color{gray!10},
    showstringspaces=false,
    columns=flexible,
    literate={https://}{https://\allowbreak}1
            {http://}{http://\allowbreak}1
            {www.}{www.\allowbreak}1
            {.com}{.com\allowbreak}1
            {.org}{.org\allowbreak}1
            {.net}{.net\allowbreak}1
            {.edu}{.edu\allowbreak}1
            {.gov}{.gov\allowbreak}1
            {/}{/\allowbreak}1
            {?}{?\allowbreak}1
            {&}{&\allowbreak}1
            {=}{=\allowbreak}1
]
I have a forecasting question that I'm interested in. Now, I want to refine and improve this question. The final output should be a fully fledged forecasting question that resolves before December 31, 2025. 

A full forecasting question includes:
- question title 
- background: A concise overview of everything a forecaster needs to know to make a forecast. 
- resolution criteria: An explanation of how exactly the question will be resolved. 

Most importantly, the question needs to be a forecasting question that is clearly and unambiguously resolvable. This means: 
1. **ALL relevant terms need to be clearly defined.**
- When defining terms, use links to e.g. Wikipedia, a dictionary, or another authoritative source, where appropriate. We need to ensure all terms are as clear as possible. 
- Be REALLY specific here. Consider, for example, the question "Will Putin and Zelenksy meet before 2026?". Since there is risk of ambiguity, what counts as a meeting definitely needs to be defined! 
- One trick that sometimes works is identifying a "pars pro toto" situation where a specific indicator is a good proxy for a larger thing. For example, "Will China and Taiwan go to war before 2027", we could ask "will more than 10 missiles within 24h be fired from either country attacking the other before 2025". We'd still have to define missile, of course. But defining "missile" is much easier than defining "war". 
2. **When December 31 comes, there ABSOLUTELY needs to be a resolution source that unambiguously informs us what the correct outcome to our question is.**
- If we already know what that source will be, that's great. In that case, we should include a link to it. Make sure the URL is not the link to some old, static file, but it should be the URL where you expect the information to be published. 
- Sometimes we don't know the resolution source in advance. In many cases that's fine, as long as we can be absolutely certain that it will exist. Consider, for example, the question "Will the current Dalai Lama die before 2026"? If this happened, there is no doubt major newspapers would report on it. In such a case, it would be fine to simply say "according to credible reporting, e.g. the NYT, Reuters, AP, ..." (again, provide examples and links). 
- One big No-no we need to avoid is relying on sources that may not exist or where it's not clear what kind of data they will produce. For example, something like "Will global investment in AI surpass $150bn according to a report published by Gartner, McKinsey, or similar" will likely NOT work for a few reasons: a) it's not clear such a report will ever exist. They might not publish a report, or they might not publish that exact number (note that pointing to recurring publications is fine! the issue is with hoping that some one-off report may exist in the future). b) MOST likely, any report like this would not exist by December 31, 2025 - these things take some time. 
3. **The question is a forecasting question**, meaning that the resolution must not yet be known as of today. It must be logically possible for the question to resolve both Yes and No. 

Let me repeat: these two things are really important and a dealbreaker. **All key terms need to be defined and there must be a resolution source come December 31.**


Next, let's talk about the background section. The background should give a clear and detailed picture of the status quo as of today. This means:
- providing all relevant context and background on the question. Basically, everything that one would need to make a first initial forecast.
- if appropriate and applicable, some info on the data and data source used to resolve the question
- the status quo needs to be clearly explained. For example, if the question is about whether X will happen, then all relevant info about X needs to be given. As of today, where is X at? What has happened recently with X? If the question is about whether some number will be higher than Y, then we MUST state what the current number is as of today (or the most recent date that data is available for). This means you will have to do some research to find out what the current number/state of the world is. 

Some more advice: 
- we're most interested in things that can realistically and plausibly happen. We therefore want to design our question such that a reasonable forecaster would likely end up with a forecast between 5% and 95%. Ideally, the final forecast should be between 30% and 70%. Pro Tip: for questions with a numerical threshold, there is a high chance you want to adapt the threshold. Aim for the number with the highest uncertainty, whether the actual number will be lower or higher (and don't feel compelled to stick to round numbers). 
- questions should be non-trivial and have high entropy. What I mean is that questions should be such that people can reasonably disagree on the outcome: Two great forecasters should be able to arrive at forecasts that are at least 20% apart, without one of them being obviously wrong. 
- questions should be difficult in the sense that doing more research (within reasonable limits) should result in a better forecast. You should avoid questions where it's hard to do better than looking up an existing number or forecast. For example, for a question like "will there be a magnitude 7 earthquake between today and December 31, 2025?" you can't do much better than just looking at base rates. Similarly for questions about e.g. stock prices - you can just look at option prices to arrive at a forecast and it's hard to do much better. Instead of those questions, we're more interested in questions where more research clearly pays off
- Often it's more interesting to ask "will event X happen" rather than "will more than Y events happen". For example, instead of asking "will the FDA approve more than X new drugs", it might be more interesting to ask whether a specific drug will get approved (note that, of course, you will have to do research to make sure the event can actually plausibly happen!).
- If you're using dates or times, make sure to always include a timezone (usually UTC). 
- Make sure to also set a start date to define the date range for eligible events. Instead of "will X happen before Y", say "will X happen between [start date] and Y". Instead of "will there be more than X before Y" say "will there be more than X between [start date] and Y". This is required to be crystal clear what events should count. 
- Avoid referencing "the next XX" when creating a question. Usually it's better to stick to "before <date>" formulations. 
- Generally, avoid questions of the form "will condition X happen before condition Y". Instead, do "will condition X happen before <DATE>".
- The question should be resolvable by a human who can find and verify the correct outcome within about 10 minutes. Keep this in mind (for example, it's preferable, although not a hard requirement to have questions with a single source of truth). Ideally, the question should focus on a single testable proposition.

Now, please return a fully-fleshed out refined forecasting question based on my original question. The question should resolve between October 15 and December 31, 2025. Faithfulness to the original question is desirable. However, your solution will be evaluated first and foremost based on quality according to the criteria above, so you should prioritise this (In particular, you should be willing to change numeric thresholds to optimise for uncertainty in the outcome). 
\end{lstlisting}
\end{tcolorbox}
\subsection{Agent Prompt to Gather Background Information for a Forecasting Question}
\label{sec:appendix-prompt-research-background}

\begin{tcolorbox}[
    breakable,
    width=\textwidth,
    left=10pt,
    right=10pt,
    top=6pt,
    bottom=6pt,
    boxrule=0.5pt,
    colback=gray!10,
    colframe=gray!50,
    title=Agent Prompt to Gather Background Information for a Forecasting Question
]
\begin{lstlisting}[
    breaklines=true,
    breakatwhitespace=true,
    basicstyle=\small\ttfamily,
    frame=single,
    framesep=2mm,
    xleftmargin=2mm,
    xrightmargin=2mm,
    backgroundcolor=\color{gray!10},
    showstringspaces=false,
    columns=flexible,
    literate={https://}{https://\allowbreak}1
            {http://}{http://\allowbreak}1
            {www.}{www.\allowbreak}1
            {.com}{.com\allowbreak}1
            {.org}{.org\allowbreak}1
            {.net}{.net\allowbreak}1
            {.edu}{.edu\allowbreak}1
            {.gov}{.gov\allowbreak}1
            {/}{/\allowbreak}1
            {?}{?\allowbreak}1
            {&}{&\allowbreak}1
            {=}{=\allowbreak}1
]
I am a professional forecaster working for a high-profile client on an important contract. I'm working on the forecasting question presented as an artifact at the end. 

Please help me gather and summarise as much information as possible about the status quo. Follow the following structure

# Background and context. 
Please collect and summarise all relevant info that someone would need to make a forecast, as well as judge the quality of the forecast. 
Where appropriate, this could be things like
- an explanation of the current status quo and context
- historical context, recent developments
- technical details
- details on the regulatory framework
- overview of stakeholders and players involved
- who are relevant candidates for a successful resolution?

Assume that readers are familiar with details from the background section of the original forecasting question (but do double-check that information is actually correct!). Try to be concise and to the point (while not omitting relevant info). You don't have to repeat things that are explained in the original background section. 

# Data and information
What data and data sources are relevant? How can I find and access the data? Please provide links

When has past data been made available? Is there a publication schedule? When and where should I expect new information to become available?

# Recent numbers and events

For questions like "will some number X be larger than Y before date Z":
- What is the current number of X? I.e., how many events have there been between the start date and now? 
- How many recent events have there been?
- What would have been the resolution the last time, e.g. what was the final number of X last year? (Ideally, we'd like to know both the total number for the full period, and a number comparable to today's number)
- Are there any relevant projections, forecasts, or base rates?

For questions like "Will X happen before Z?" or "Will the outcome of X be Y?":
- Has X happened before? If so, when and how often?
- What was the outcome of X earlier?
- Are there any relevant projections, forecasts, or base rates?


# Trends
Please list factors and trends that make a YES or a NO resolution more likely. 

Please reply in the following format: 
<reply format>
# Background and context
[summary of everything you deem relevant]

# Data and information
## data sources
[relevant data sources and how to access them]

## data timing
[information about when information will likely become available]

# Recent numbers and events
## Headline numbers/events
[summary of recent headline numbers and events]

## Other related numbers/events
[...]

# Trends
## In favour of YES
[factors that make a YES resolution more likely]

## In favour of NO
[factors that make a NO resolution more likely]
</reply format>

It is VERY important that you add links and sources, and explain where you found your information. This is important work, and any mistakes could be hugely embarrassing. 

When writing your reply, please try to make it such that the information is easily digestible for a human reader: Be concise, structure your replies in a way that make sense, explain why things you mention are important, give the reader a picture that will help them to understand what's going on and why that matters in the context of the forecasting question. Again: this should be for a human reader who quickly wants to understand what's going on. 

\end{lstlisting}
\end{tcolorbox}

\subsection{Question Evaluation Prompts}
\label{sec:appendix-question-evaluation-prompts}

\subsection{Agent Prompt to Assess Whether Question Is a Meaningful Forecasting Question}
\label{sec:appendix-prompt-assess-question-is-good}

\begin{tcolorbox}[
    breakable,
    width=\textwidth,
    left=10pt,
    right=10pt,
    top=6pt,
    bottom=6pt,
    boxrule=0.5pt,
    colback=gray!10,
    colframe=gray!50,
    title=Agent Prompt to Assess Whether Question Is a Meaningful Forecasting Question
]
\begin{lstlisting}[
    breaklines=true,
    breakatwhitespace=true,
    basicstyle=\small\ttfamily,
    frame=single,
    framesep=2mm,
    xleftmargin=2mm,
    xrightmargin=2mm,
    backgroundcolor=\color{gray!10},
    showstringspaces=false,
    columns=flexible,
    literate={https://}{https://\allowbreak}1
            {http://}{http://\allowbreak}1
            {www.}{www.\allowbreak}1
            {.com}{.com\allowbreak}1
            {.org}{.org\allowbreak}1
            {.net}{.net\allowbreak}1
            {.edu}{.edu\allowbreak}1
            {.gov}{.gov\allowbreak}1
            {/}{/\allowbreak}1
            {?}{?\allowbreak}1
            {&}{&\allowbreak}1
            {=}{=\allowbreak}1
]
I am an admin at Metaculus, the forecasting platform, and I'm gathering questions for an upcoming forecasting tournament. Please help me assess the quality of a forecasting question (represented by title, resolution criteria, and background - provided as an artifact below).

## Task

The main question I would like you to focus on is "is this a good forecasting question?".

A good forecasting question

1. is somewhat difficult, in the sense that if you do more research, you should expect to arrive at a different forecast (a bad question is: "will this coin come up heads?" - there is not much to do except forecasting the base rate. An ok question is something like "will the price of BTC reach X in 2025"? - you'd need to look up BTC option prices and derive the probability from them, so there is some work involved. But once you did that, you shouldn't expect to move away much from that forecast. A good question would be "will candidate X win in this contested election" - you'd have to a lot of work to interpret trends, polling results etc.)
2. has "high entropy":
    1. The question should be non-trivial, with an answer that's not almost certainly true or false
    2. We should expect to learn something between now and when the question resolves. 
        1. If the question asks about whether an event will occur or not, "nothing happened" of course qualifies as an update
        2. if something definitely happened in the real world, but there is a high risk that we won't know what's going on because the relevant information might not be updated or published until after the resolution date then that's bad
    3. There is room for disagreement: For example, there be two good forecasters whose forecasts reasonably differ by a large amount (e.g. at least 20 percentage points) without someone being clearly wrong a priori. 

## Reply

rationale_quality:

Your final reply should start with a rationale. In this rationale, you should explicitly reason about every point in the list above. In addition, you should write about any other considerations that come to mind. Lastly, you should have a section  "final answer reasoning" where you reason about your final answer on whether this is a good question. 

final_answer_quality:

You should answer with one of the three following options, nothing else: 

- "bad" - if you think the question is bad (for example, because it's trivial, or we will see data issues with high probability, or for other reasons)
- "meh" - if you think the question is meh. Not terrible, but not quite what you would present to forecasters in a public forecasting tournament. 
- "good" - if you think this is good to use for a public forecasting tournament. However, it might have some issues or not offer a very rich forecasting experience
- "great" - if you think this is a great question for a public forecasting tournament. 

\end{lstlisting}
\end{tcolorbox}
\subsection{Agent Prompt to Assess Whether Question Can Be Clearly and Unambiguously Resolved}
\label{sec:appendix-prompt-assess-ambiguity}

\begin{tcolorbox}[
    breakable,
    width=\textwidth,
    left=10pt,
    right=10pt,
    top=6pt,
    bottom=6pt,
    boxrule=0.5pt,
    colback=gray!10,
    colframe=gray!50,
    title=Agent Prompt to Assess Whether Question Can Be Clearly and Unambiguously Resolved
]
\begin{lstlisting}[
    breaklines=true,
    breakatwhitespace=true,
    basicstyle=\small\ttfamily,
    frame=single,
    framesep=2mm,
    xleftmargin=2mm,
    xrightmargin=2mm,
    backgroundcolor=\color{gray!10},
    showstringspaces=false,
    columns=flexible,
    literate={https://}{https://\allowbreak}1
            {http://}{http://\allowbreak}1
            {www.}{www.\allowbreak}1
            {.com}{.com\allowbreak}1
            {.org}{.org\allowbreak}1
            {.net}{.net\allowbreak}1
            {.edu}{.edu\allowbreak}1
            {.gov}{.gov\allowbreak}1
            {/}{/\allowbreak}1
            {?}{?\allowbreak}1
            {&}{&\allowbreak}1
            {=}{=\allowbreak}1
]
I am an admin at Metaculus, the forecasting platform and I'm gathering questions for an upcoming forecasting tournament. Please help me assess the quality of a forecasting question (represented by title, resolution criteria, and background - provided as an artifact below).

## Task

The main criterion I would like you to focus on is this: Do you think the question can be clearly and unambiguously resolved? 

To that effect, here is a list of subquestions you should consider:

1. Are all key terms well-defined, providing either a short definition or a link to an authoritative site (e.g. Wikipedia, a dictionary, an official organization, etc.)? Is there no definitional ambiguity that would make a reasonable person uncertain about what could be meant?
2. Is the resolution time/date unambiguous? Every date/time should have a timezone specified and a year.
3. If any numeric cutoffs are mentioned, are all thresholds and cutoffs in the question explicitly defined (e.g., >= 50% vs >50%)?
4. Is the question robust against unexpected technicalities or gotchas? Or is there anything that we should pay attention to that could make the question resolve contrary to the spirit of the question? 
5. Will the content of the resolution source be unambiguous? Some examples:
    - if the resolution source reports a number, that's unambiguous. If it's not exactly clear what the relevant number is (if there may be multiple) then that's not unambiguous.
    - If the question asks whether a newspaper will write X then that's likely unambiguous. If we have to interpret something the newspaper writes (e.g. what does it mean for someone to "endorse" something) then that's likely ambiguous.
    Basically, what I want to know: If 10 people look at the question and the source, will they all broadly agree on the outcome? Or is there a high chance for disagreements, subjectivity etc? For this subquestion, please provide a value between 0 and 100 as an answer. 0 indicates "source is very ambiguous, there will likely be more than one reasonable interpretation of the source", 100 means "everything is crystal clear, no two informed and well-intentioned people would reasonably disagree.
    

## Reply

Your final reply should have the following parts:

1. rationale
    1. answers to the checklist
    2. additional comments/feedback
    3. rationale for the final answer to the question whether we will be able to clearly and unambiguously resolve the question
2. final answer

rationale_ambiguity:
Please start your answer by going through the checklist above. Mirroring the structure of the original checklist, for each item, please provide an answer as well as an explanation for your answer. For the first 4 items and subitems, the answer should be "True", "False" or "Does not apply". For the 5th item, the answer should be a number between 0 and 100.

Reply with the following structure:

**checklist**

1. [answer] - explanation
2. [answer] - explanation
3. ...
4. ...
5. ...

After you've gone through the list, you may write a section **additional comments** with additional explanations or feedback for the question creator, if you like.

Next, write a session **final answer reasoning**. This section should include the reasoning for your assessment about whether or not this question can be clearly and unambiguously resolved. Are the terms clear? Is it likely that two reasonable people would disagree on the outcome?

final_answer_ambiguity:
Please answer with one of the following options, just a single word, nothing else: 
- "bad" - if you think the question is bad (for example, because there is a high risk of ambiguities or disputes)
- "meh" - if you think the question is meh. Not terrible, but not quite what you would present to forecasters in a public forecasting tournament. 
- "good" - if you think this is good to use for a public forecasting tournament. It might have some issues, but those are minor and most likely won't lead to a situation where people would reasonably disagree on the resolution
- "great" - if you think this is a great question for a public forecasting tournament that is crystal clear and unambiguous.

\end{lstlisting}
\end{tcolorbox}
\subsection{Agent Prompt to Assess Whether Question Can Be Autonomously Resolved by an AI Agent}
\label{sec:appendix-prompt-assess-automated-resolvability}

\begin{tcolorbox}[
    breakable,
    width=\textwidth,
    left=10pt,
    right=10pt,
    top=6pt,
    bottom=6pt,
    boxrule=0.5pt,
    colback=gray!10,
    colframe=gray!50,
    title=Agent Prompt to Assess Whether Question Can Be Autonomously Resolved by an AI Agent
]
\begin{lstlisting}[
    breaklines=true,
    breakatwhitespace=true,
    basicstyle=\small\ttfamily,
    frame=single,
    framesep=2mm,
    xleftmargin=2mm,
    xrightmargin=2mm,
    backgroundcolor=\color{gray!10},
    showstringspaces=false,
    columns=flexible,
    literate={https://}{https://\allowbreak}1
            {http://}{http://\allowbreak}1
            {www.}{www.\allowbreak}1
            {.com}{.com\allowbreak}1
            {.org}{.org\allowbreak}1
            {.net}{.net\allowbreak}1
            {.edu}{.edu\allowbreak}1
            {.gov}{.gov\allowbreak}1
            {/}{/\allowbreak}1
            {?}{?\allowbreak}1
            {&}{&\allowbreak}1
            {=}{=\allowbreak}1
]
I am an admin at Metaculus, the forecasting platform and I'm gathering questions for an upcoming forecasting tournament. Please help me assess the quality of a forecasting question (represented by title, resolution criteria, and background, and some additional information - provided as an artifact below). 

## Task 

The main criterion I would like you to focus on is this: Do you think the question can be clearly and unambiguously resolved by an AI agent? 

To that effect, here is a list of subquestions you should consider:

1. Will it be trivially possible to locate the resolution source? This is the case if
    1. a resolution source is clearly and unambiguously specified such that it can be identified and located without further clarification
    2. the resolution source isn't fully specified in advance, but this does not pose a problem and it will be trivial to identify a suitable resolution source. (For example, consider a question about whether the Dalai Lama will die before date X. Even if the resolution criteria only say "the resolution source will be credible reporting, e.g. from the NYT, reuters, or AP, or similar", it will still be trivial to verify that the event has indeed happened.)
2. Will the resolution source exist with 99% probability at the time of the resolution? (Don't get too hung up on the 99% number. Basically, what I want to know: Can I be very certain that the source will exist when I try checking it when the question resolves? For example, this is usually the case for recurring publications. However, we can't rely on a certain one-off report being reproduced in the future.)
3. If the source is already specified and known in advance: 
    1. Does the source actually exist (or will surely exist)? (only answer after having actually checked and accessed the source)
    2. Is it freely accessible? 
    3. If the question names a specific section/column/variable, does that section/column/variable actually exist (or can we be sure that it will exist at the time the question resolves)? Has the methodology of that variable changed recently in a way that's not reflected by the resolution criteria? Is the variable subject to change between now and time when the question resolves?
4. Can you access the source and see what it says at the moment? If there is some number or value you should be able to see, can you actually see it? Report the current value in your explanation, if appropriate. If you can't access the value, why not? Are you sure this is just a temporary issue? Use classic rationalist techniques such as "noticing your confusion" to think about this. 
5. The question could be resolved by a human within 10 minutes, given the provided information. 
    
## Reply

Your final reply should have the following parts. 
1. rationale 
    1. answers to the checklist
    2. additional comments/feedback
    3. rationale for the final answer to the question whether an AI agent will be able to resolve the question automatically
2. final answer 

rationale_resolvability: 
Please start your answer by going through the checklist above. Mirroring the structure of the original checklist, for each item, please provide an answer as well as an explanation for your answer. For each item and subitem, the answer should be "True", "False" or "Does not apply". 

Reply with the following structure:

**checklist**

1. [answer] - explanation
    1. [answer] - explanation
    2. [answer] - explanation
2. [answer] - explanation
3. ...
    1. ...
    2. ...
    3. ...
    4. ...

Again, make sure to only answer after you've done everything necessary to convince yourself that the source will be available and you can access it.

After you've gone through the list, you may write a section **additional comments** with additional explanations or feedback for the question creator, if you like. 

Next, write a session **final answer reasoning**. This section should include the reasoning for your estimate on whether or not an AI agent will be able to resolve the question clearly and unambiguously on its own.

final_answer_resolvability: 
Please provide an estimate of whether or not an AI agent will be able to resolve the question. The agent is given the same tools you currently have at your disposal and a similar budget. That means, for example, that if the source should currently exist, but you can't access it, then chances are a future agent will also not be able to access it. Your estimate should reflect the probability that an agent will be able to provide a clear and unambiguous resolution. 
Note: In the checklist I asked about whether a human would be able to resolve the question within 10 minutes. If you think that a human won't be able to do this, then this should be significant evidence that an agent may not be able to do this with >95% probability. 

Answer with one of the following options, nothing else:
- "very certainly no" (if you think this will be very hard with <5% success probability)
- "probably no" - (if you think it would fail more likely than not, success probability <50%)
- "probably yes" - (if you think it might work, but there is a decent chance it might fail, success probability between 50% and 90%)
- "very certainly yes" - if you're confident an agent just like yourself would be able to resolve the question with a success probability >95%). 

\end{lstlisting}
\end{tcolorbox}
\subsection{Agent Prompt to Assess a Forecasting Question via Forecasting}
\label{sec:appendix-prompt-assess-question-via-forecasting}

\begin{tcolorbox}[
    breakable,
    width=\textwidth,
    left=10pt,
    right=10pt,
    top=6pt,
    bottom=6pt,
    boxrule=0.5pt,
    colback=gray!10,
    colframe=gray!50,
    title=Agent Prompt to Assess a Forecasting Question via Forecasting
]
\begin{lstlisting}[
    breaklines=true,
    breakatwhitespace=true,
    basicstyle=\small\ttfamily,
    frame=single,
    framesep=2mm,
    xleftmargin=2mm,
    xrightmargin=2mm,
    backgroundcolor=\color{gray!10},
    showstringspaces=false,
    columns=flexible,
    literate={https://}{https://\allowbreak}1
            {http://}{http://\allowbreak}1
            {www.}{www.\allowbreak}1
            {.com}{.com\allowbreak}1
            {.org}{.org\allowbreak}1
            {.net}{.net\allowbreak}1
            {.edu}{.edu\allowbreak}1
            {.gov}{.gov\allowbreak}1
            {/}{/\allowbreak}1
            {?}{?\allowbreak}1
            {&}{&\allowbreak}1
            {=}{=\allowbreak}1
]
You are a professional forecaster interviewing for a job.

You are tasked with giving a probabilistic forecast for the question artifact shown at the end.

Here are top tips from good forecasters:

If an event was anticipated to happen in a certain timeframe, but 80% of that time has passed and there is no recent news or updates about it happening soon, then you should be skeptical that it will happen on the originally stated timeframe. It probably means that it will be delayed or plans have changed. Some examples:
If Elon Musk says definitely that robotaxis will be available in 1 year, but after 11 months there have not been any specific public updates confirming that robotaxis will be available then it is highly unlikely to happen in the remaining month.
If Donald Trump says that he will definitely have a deal signed in 90 days, but 70 days have passed without updates on the progress, then it is highly unlikely there will be a deal in the remaining 20 days.

Think about base rates for similar events in the past. Sometimes a base rate is the best you can do, if you can't find much information about the question. Example questions where the base rate is a particularly good starting point include:
Will the temperature in Miami exceed 100 degrees next month?
Will there be an earthquake of magnitude 5 or more in San Francisco in 2026?

Sometimes finding a good base rate is difficult, especially when the events are relatively unique. In these cases, you need to put more weight on the "inside view" which means weighing considerations that appear specific to the situation that lead to a very different forecast than base rates would dictate. You will need to use your own judgment.

Put extra weight on the status quo outcome since the world changes slowly most of the time. This is especially true when coordination or agreement between people or organizations is required. For example, signing multi-national treaties and passing legislation often take longer than one might imagine from reading the news.

Think about if there are seasonal effects. For example, the sales of homes or travel are likely to have seasonal patterns.

Think about what the current trend is and if it makes sense to extrapolate, or not. Some things like stock prices are effectively random walks, so recent trends likely don't matter. Other trends have momentum, like the number of COVID cases from day to day.

Think about the scope of the question.

Think about the incentives and power of any influential people involved in the situation. For example, Putin has the power to single-handedly dictate Russian military or diplomatic response.

Sometimes there are multiple data sources for the same number that have very different values. For example, Trading Economics reports 1.84B UAH for Ukrainian debt while the IMF reports a value of 7B UAH. These are both reputable sources, but they are using different definitions in their reporting. It is important to focus on the source used in the resolution criteria.

(10) Pre-mortem. Think about how you are most likely to be wrong. Imagine that you are writing a letter to your future self that you will open and read once the outcome is known. In the letter you try to explain to your future self the most likely way that your forecast will be deemed to be a poor forecast. Are you most worried about missing a key piece of information? What is the biggest uncertainty and would keep you awake at night?

(11) More general advice:
    - Pay close attention to the exact wording and resolution source in the resolution criteria. Sometimes newspaper articles will cite a number that is significantly different from the number in the resolution criteria. Make sure to pay attention to the resolution criteria.
    - Like a good forecaster, you should use your own judgment to come to the most accurate forecast!

Before answering, please write a detailed rationale that explains your reasoning. As part of that, also please include:
(a) The time left until the outcome to the question is known.
(b) The status quo outcome if nothing changed.
(c) Think about answering the question with different scopes to help ensure that you have a self consistent view and have considered the broader context.
    For example, imagine the question is: Will a company declare bankruptcy in the next 3 months? It can be useful to force yourself to forecast the probability of bankruptcy over the next 1 year, 2 years, and 5 years. Doing this in a self-consistent way helps to force you to consider the scope explicitly.
    If you forecast 40% in 12 months, then you might forecast 10% in 3 months to be scope sensitive. However, it is also possible that the bankruptcy risk is higher in the near term, so it could still be 25%.
    You will need to use your judgment. Being scope insensitive is a common cognitive bias and this exercise is meant to help combat this bias by forcing you to explicitly consider the question's scope.
(d) A brief description of a scenario that results in a No outcome.
(e) A brief description of a scenario that results in a Yes outcome.

The last thing you write is your final answer. This should be the probability you assign to a "YES" outcome, written as a percentage, i.e., a number between 0 and 100.

So again, the answer format should be:
rationale_forecast (a detailed explanation of your reasoning, which should also include the points mentioned above)
final_answer_forecast (your probability)

[Forecasting question title]
[Background]
[Resolution criteria]
[Additional information]
\end{lstlisting}
\end{tcolorbox}

\subsection{Forecasting Prompts}
\label{sec:appendix-forecasting-prompts}

\subsection{Agent Prompt to Obtain Research for a Forecast on a Question}
\label{sec:appendix-prompt-final-forecast-research}

\begin{tcolorbox}[
    breakable,
    width=\textwidth,
    left=10pt,
    right=10pt,
    top=6pt,
    bottom=6pt,
    boxrule=0.5pt,
    colback=gray!10,
    colframe=gray!50,
    title=Agent Prompt to Obtain Research for a Forecast on a Question
]
\begin{lstlisting}[
    breaklines=true,
    breakatwhitespace=true,
    basicstyle=\small\ttfamily,
    frame=single,
    framesep=2mm,
    xleftmargin=2mm,
    xrightmargin=2mm,
    backgroundcolor=\color{gray!10},
    showstringspaces=false,
    columns=flexible,
    literate={https://}{https://\allowbreak}1
            {http://}{http://\allowbreak}1
            {www.}{www.\allowbreak}1
            {.com}{.com\allowbreak}1
            {.org}{.org\allowbreak}1
            {.net}{.net\allowbreak}1
            {.edu}{.edu\allowbreak}1
            {.gov}{.gov\allowbreak}1
            {/}{/\allowbreak}1
            {?}{?\allowbreak}1
            {&}{&\allowbreak}1
            {=}{=\allowbreak}1
]
You are a professional forecaster applying for a job with a high-profile client. As part of a work trial, you're working on the forecasting question presented as an artifact at the end. This is a VERY important task, so it's important to do it as well as possible. 

The plan is to attack the problem in a 2-step process. 
Step 1 is collecting and summarising all relevant information. 
Step 2 is creating a probabilistic forecast based on that information. (no internet connection!)

You are at step 1. Your goal is to understand the topic as thoroughly as possible and find every piece of information that could be useful for the final forecast. 

Please collect and summarise all relevant info that someone would need to make a forecast. You will find some background info at the end. Once you're at the forecasting step, you will no longer have access to the internet or be able to use any other kinds of tools, so you need to make sure that you have everything you need for the forecast. This could be things like the following:

<suggested things to think about>
# Background and context. 
- an explanation of the current status quo and context
- historical context, recent developments
- technical details
- details on the regulatory framework
- overview of stakeholders and players involved
- When and where should we expect new information to become available?

# Recent numbers and events
For questions like "will some number X be larger than Y before date Z":
- What is the current number of X? I.e., how many events have there been between the start date and now? 
- How many recent events have there been?
- What would have been the resolution the last time, e.g. what was the final number of X last year? (Ideally, we'd like to know both the total number for the full period, and a number comparable to today's number)
- Are there any relevant projections, forecasts, or base rates?

For questions like "Will X happen before Z?" or "Will the outcome of X be Y?":
- Has X happened before? If so, when and how often?
- What was the outcome of X earlier?
- Are there any relevant projections, forecasts, or base rates?

# Trends and arguments for Yes/No
Please list factors and trends that make a YES or a NO resolution more likely. 
</suggested things to think about>

It may also help you to think from the perspective of a forecaster to guide your research. Here are some general forecasting tips.  I'm providing these so you have some background on what you'll be asked to do in the next step. 
<forecasting tips>
Think about base rates for similar events in the past. Sometimes finding a good base rate is difficult, especially when the events are relatively unique. In these cases, you need to put more weight on the "inside view" which means weighing considerations that appear specific to the situation that lead to a very different forecast than base rates would dictate. 

Put extra weight on the status quo outcome since the world changes slowly most of the time. This is especially true when coordination or agreement between people or organizations is required. 

Think about if there are seasonal effects. For example, the sales of homes or travel are likely to have seasonal patterns.

Think about what the current trend is and if it makes sense to extrapolate, or not. Some things like stock prices are effectively random walks, so recent trends likely don't matter. Other trends have momentum, like the number of COVID cases from day to day.

Think about the scope of the question.

Think about the incentives and power of any influential people involved in the situation. For example, Putin has the power to single-handedly dictate Russian military or diplomatic response.

Sometimes there are multiple data sources for the same number that have very different values. For example, Trading Economics reports 1.84B UAH for Ukrainian debt while the IMF reports a value of 7B UAH. These are both reputable sources, but they are using different definitions in their reporting. It is important to focus on the source used in the resolution criteria.

Pre-mortem. Think about how you are most likely to be wrong. Imagine that you are writing a letter to your future self that you will open and read once the outcome is known. In the letter you try to explain to your future self the most likely way that your forecast will be deemed to be a poor forecast. Are you most worried about missing a key piece of information? What is the biggest uncertainty and would keep you awake at night?
</forecasting tips>

Remember: Your task is to provide all information that you may need and find helpful when making a forecast in the next step. When crafting your reply, I suggest you follow a structure that mimics the topics suggested above: 
1. Background
2. Recent events
3. Trends and arguments for Yes/No

Make sure to present and structure the information in a way that will make it easiest for your future self to use it for a forecast. For example, it may be helpful to provide your context and reasoning, as well as your own level of uncertainty, alongside any factual claims you report. 

[Forecasting question title]
[Background]
[Resolution criteria]
[Additional information]
\end{lstlisting}
\end{tcolorbox}
\subsection{Agent Prompt to Give a Probabilistic Forecast for a Question}
\label{sec:appendix-prompt-final-forecast-probability}

\begin{tcolorbox}[
    breakable,
    width=\textwidth,
    left=10pt,
    right=10pt,
    top=6pt,
    bottom=6pt,
    boxrule=0.5pt,
    colback=gray!10,
    colframe=gray!50,
    title=Agent Prompt to Give a Probabilistic Forecast for a Question
]
\begin{lstlisting}[
    breaklines=true,
    breakatwhitespace=true,
    basicstyle=\small\ttfamily,
    frame=single,
    framesep=2mm,
    xleftmargin=2mm,
    xrightmargin=2mm,
    backgroundcolor=\color{gray!10},
    showstringspaces=false,
    columns=flexible,
    literate={https://}{https://\allowbreak}1
            {http://}{http://\allowbreak}1
            {www.}{www.\allowbreak}1
            {.com}{.com\allowbreak}1
            {.org}{.org\allowbreak}1
            {.net}{.net\allowbreak}1
            {.edu}{.edu\allowbreak}1
            {.gov}{.gov\allowbreak}1
            {/}{/\allowbreak}1
            {?}{?\allowbreak}1
            {&}{&\allowbreak}1
            {=}{=\allowbreak}1
]
You are a professional forecaster applying for a job with a high-profile client. As part of a work trial, you're working on the forecasting question presented as an artifact at the end. This is a VERY important task, so it's important to do it as well as possible. 

The plan is to attack the problem in a 2-step process. 
Step 1 is collecting and summarising all relevant information. 
Step 2 is creating a probabilistic forecast based on that information

You are at step 2. Based on the given information, your job is to provide the best possible probabilistic forecast. Your answer will be judged using a proper scoring rule, so it is important that you really report your true belief. 

Here are some general forecasting tips.  
<forecasting tips>
If an event was anticipated to happen in a certain timeframe, but 80% of that time has passed and there is no recent news or updates about it happening soon, then you should be more skeptical that it will happen on the originally stated timeframe. It probably means that it will be delayed or plans have changed. Some examples:

Think about base rates for similar events in the past. Sometimes finding a good base rate is difficult, especially when the events are relatively unique. In these cases, you need to put more weight on the "inside view" which means weighing considerations that appear specific to the situation that lead to a very different forecast than base rates would dictate. 

Put extra weight on the status quo outcome since the world changes slowly most of the time. This is especially true when coordination or agreement between people or organizations is required. 

Think about if there are seasonal effects. For example, the sales of homes or travel are likely to have seasonal patterns.

Think about what the current trend is and if it makes sense to extrapolate, or not. Some things like stock prices are effectively random walks, so recent trends likely don't matter. Other trends have momentum, like the number of COVID cases from day to day.

Think about the scope of the question.

Think about the incentives and power of any influential people involved in the situation. For example, Putin has the power to single-handedly dictate Russian military or diplomatic response.

Sometimes there are multiple data sources for the same number that have very different values. For example, Trading Economics reports 1.84B UAH for Ukrainian debt while the IMF reports a value of 7B UAH. These are both reputable sources, but they are using different definitions in their reporting. It is important to focus on the source used in the resolution criteria.

Pre-mortem. Think about how you are most likely to be wrong. Imagine that you are writing a letter to your future self that you will open and read once the outcome is known. In the letter you try to explain to your future self the most likely way that your forecast will be deemed to be a poor forecast. Are you most worried about missing a key piece of information? What is the biggest uncertainty and would keep you awake at night?

Don't anchor to round numbers. It's ok for probabilities to have odd numbers. 

Pay close attention to the exact wording and resolution source in the resolution criteria.

You should reason both in probability space (e.g., "an event has a 5% chance of happening") as well as in odds space (e.g. "an event has a 1 in 20 chance of happening"). This helps with gaining an intuitive and self-consistent sense of the probabilities involved. 

When reasoning about probabilities, think about them in the context of bets. Would you be willing to pay 90 cents for a coupon that would be out $1 if the event happens? Conversely, would you prefer to pay 10 cents to take the other side of the bet? Only if you're indifferent between the two sides have you arrived at a consistent probability. 
</forecasting tips>

Before answering, use your own internal thinking extensively. Then, write a brief summary which includes at least the following items:

(a) The time left until the outcome to the question is known.
(b) The status quo outcome if nothing changed.
(c) Think about answering the question with different scopes to help ensure that you have a self consistent view and have considered the broader context.
    For example, imagine the question is: Will a company declare bankruptcy in the next 3 months? It can be useful to force yourself to forecast the probability of bankruptcy over the next 1 year, 2 years, and 5 years. Doing this in a self-consistent way helps to force you to consider the scope explicitly.
    If you forecast 40% in 12 months, then you might forecast 10% in 3 months to be scope sensitive. However, it is also possible that the bankruptcy risk is higher in the near term, so it could still be 25%.
    You will need to use your judgment. Being scope insensitive is a common cognitive bias and this exercise is meant to help combat this bias by forcing you to explicitly consider the question's scope.
(d) A brief reasoning for why a No outcome might be likely. 
(e) A brief reasoning for why a Yes outcome.
(f) Reason about your probabilities in terms of bets and make sure that you'd be indifferent between both sides. 

The last thing you write is your final answer. This should be the probability you assign to a "YES" outcome, written as a percentage, i.e., a number between 0 and 100.

[Forecasting question title]
[Background]
[Resolution criteria]
[Additional information]
[Research summary]
\end{lstlisting}
\end{tcolorbox}
\subsection{Agent Prompt to Decompose a Question into Subquestions}
\label{sec:appendix-prompt-subquestion-decomposition}

\begin{tcolorbox}[
    breakable,
    width=\textwidth,
    left=10pt,
    right=10pt,
    top=6pt,
    bottom=6pt,
    boxrule=0.5pt,
    colback=gray!10,
    colframe=gray!50,
    title=Agent Prompt to Decompose a Question into Subquestions
]
\begin{lstlisting}[
    breaklines=true,
    breakatwhitespace=true,
    basicstyle=\small\ttfamily,
    frame=single,
    framesep=2mm,
    xleftmargin=2mm,
    xrightmargin=2mm,
    backgroundcolor=\color{gray!10},
    showstringspaces=false,
    columns=flexible,
    literate={https://}{https://\allowbreak}1
            {http://}{http://\allowbreak}1
            {www.}{www.\allowbreak}1
            {.com}{.com\allowbreak}1
            {.org}{.org\allowbreak}1
            {.net}{.net\allowbreak}1
            {.edu}{.edu\allowbreak}1
            {.gov}{.gov\allowbreak}1
            {/}{/\allowbreak}1
            {?}{?\allowbreak}1
            {&}{&\allowbreak}1
            {=}{=\allowbreak}1
]
You are a professional forecaster working for a high-profile client as part of a larger forecasting team. You're working on the forecasting question presented as an artifact at the end. This is a VERY important task, so it's important to do it as well as possible.

The plan is to attack the problem in a multi-step process.
Step 1 is decomposing the question into several subquestions.
Step 2 is making a forecast on each of these subquestions. This will be done by your colleagues, who will then give the final forecast back to you. Note: they don't know the top-level question!
Step 3 is making a forecast on the top-level question, while using the forecasts for the subquestions as additional input (you'll have both the original question and the subquestions)

You're at step 1. Your task, for now, is therefore to come up with between 1 and 5 subquestions that you think would be useful for creating the final forecast later on.

Subquestions should have
- question title: formulated as a question
- background: A concise overview of everything a forecaster needs to know to make a forecast.
- resolution criteria: An explanation of how exactly the question will be resolved.

Some tips on how to think about your subquestions:

<subquestion tips>
- Subquestions should be in the service of the final forecast. Doing some research and creating a forecast on the subquestion should help you improve the final forecast
- Subquestions need to be clear, unambiguous, and resolvable. This means: it has to be crystal clear what a forecaster is asked to do. The questions will be given to your colleagues, who will not be able to talk to you. Therefore, it needs to be SUPER clear what exactly they should forecast on.
- **ALL relevant terms need to be clearly defined.**
- Subquestions should be diverse and orthogonal. There is no point in asking similar questions, and you should be mindful of your colleagues' time.
- Again, look for subquestions that are meaningful and different. Don't just ask for variations of the same original question.
- It's completely ok to suggest fewer than 5 subquestions!
- Subquestions ought to have "high entropy", i.e., they should be non-trivial questions, with answers that are not almost certainly true or false. You should learn something important from them.
- Subquestions MUST be phrased independently of the top-level question. Your colleagues, who will make the forecasts on the subquestions, do NOT know the top-level question and don't know any details except what you give them.
</subquestion tips>

You might also want to keep in mind a few general tips for forecasters. These will be important for you when you reach step 3 and will also be what your colleagues have in mind when doing their forecasts.

<general forecasting tips>
Think about base rates for similar events in the past. Sometimes finding a good base rate is difficult, especially when the events are relatively unique. In these cases, you need to put more weight on the "inside view" which means weighing considerations that appear specific to the situation that lead to a very different forecast than base rates would dictate.

Put extra weight on the status quo outcome since the world changes slowly most of the time. This is especially true when coordination or agreement between people or organizations is required.

Think about if there are seasonal effects. For example, the sales of homes or travel are likely to have seasonal patterns.

Think about what the current trend is and if it makes sense to extrapolate, or not. Some things like stock prices are effectively random walks, so recent trends likely don't matter. Other trends have momentum, like the number of COVID cases from day to day.

Think about the scope of the question. How would your answer change if the timeframe were smaller/bigger? How would your answer change if the numbers involved were smaller/bigger?

Think about the incentives and power of any influential people involved in the situation. For example, Putin has the power to single-handedly dictate Russian military or diplomatic response.

Sometimes there are multiple data sources for the same number that have very different values. For example, Trading Economics reports 1.84B UAH for Ukrainian debt while the IMF reports a value of 7B UAH. These are both reputable sources, but they are using different definitions in their reporting. It is important to focus on the source used in the resolution criteria.

Pre-mortem. Think about how you are most likely to be wrong. Imagine that you are writing a letter to your future self that you will open and read once the outcome is known. In the letter you try to explain to your future self the most likely way that your forecast will be deemed to be a poor forecast. Are you most worried about missing a key piece of information? What is the biggest uncertainty and would keep you awake at night?
</general forecasting tips>

Those are just some general tips and reminders about how forecasters tend to think. Again, your task is now to come up with a set of 1 to 5 subquestions to help you create the best possible forecast in step 3. Remember, your subquestions need to be completely independent of the top-level question.
Think internally before you write. In particular, think about the best possible set of questions you can ask.
\end{lstlisting}
\end{tcolorbox}

\subsubsection{Agent Prompt to Forecast Using Subquestion Research}
\label{sec:appendix-prompt-subquestion-forecast}

\begin{tcolorbox}[
    breakable,
    width=\textwidth,
    left=10pt,
    right=10pt,
    top=6pt,
    bottom=6pt,
    boxrule=0.5pt,
    colback=gray!10,
    colframe=gray!50,
    title=Agent Prompt to Forecast Using Subquestion Research
]
\begin{lstlisting}[
    breaklines=true,
    breakatwhitespace=true,
    basicstyle=\small\ttfamily,
    frame=single,
    framesep=2mm,
    xleftmargin=2mm,
    xrightmargin=2mm,
    backgroundcolor=\color{gray!10},
    showstringspaces=false,
    columns=flexible,
    literate={https://}{https://\allowbreak}1
            {http://}{http://\allowbreak}1
            {www.}{www.\allowbreak}1
            {.com}{.com\allowbreak}1
            {.org}{.org\allowbreak}1
            {.net}{.net\allowbreak}1
            {.edu}{.edu\allowbreak}1
            {.gov}{.gov\allowbreak}1
            {/}{/\allowbreak}1
            {?}{?\allowbreak}1
            {&}{&\allowbreak}1
            {=}{=\allowbreak}1
]
You are a professional forecaster applying for a job with a high-profile client. As part of a work trial, you're working on the forecasting question presented as an artifact at the end. This is a VERY important task, so it's important to do it as well as possible.

The plan is to attack the problem in a 3-step process.
Step 1 is collecting and summarising all relevant information.
Step 2 is generating forecasts that are relevant to the question.
Step 3 is creating a probabilistic forecast based on that information

You'll find the results from step 2 in the "subforecasts" artifact. Please use these if you find them helpful, but be careful not to anchor to them too much.

You are at step 3. Based on the given information, your job is to provide the best possible probabilistic forecast. Your answer will be judged using a proper scoring rule, so it is important that you really report your true belief.

Here are some general forecasting tips.
<forecasting tips>
If an event was anticipated to happen in a certain timeframe, but 80% of that time has passed and there is no recent news or updates about it happening soon, then you should be more skeptical that it will happen on the originally stated timeframe. It probably means that it will be delayed or plans have changed. Some examples:

Think about base rates for similar events in the past. Sometimes finding a good base rate is difficult, especially when the events are relatively unique. In these cases, you need to put more weight on the "inside view" which means weighing considerations that appear specific to the situation that lead to a very different forecast than base rates would dictate.

Put extra weight on the status quo outcome since the world changes slowly most of the time. This is especially true when coordination or agreement between people or organizations is required.

Think about if there are seasonal effects. For example, the sales of homes or travel are likely to have seasonal patterns.

Think about what the current trend is and if it makes sense to extrapolate, or not. Some things like stock prices are effectively random walks, so recent trends likely don't matter. Other trends have momentum, like the number of COVID cases from day to day.

Think about the scope of the question.

Think about the incentives and power of any influential people involved in the situation. For example, Putin has the power to single-handedly dictate Russian military or diplomatic response.

Sometimes there are multiple data sources for the same number that have very different values. For example, Trading Economics reports 1.84B UAH for Ukrainian debt while the IMF reports a value of 7B UAH. These are both reputable sources, but they are using different definitions in their reporting. It is important to focus on the source used in the resolution criteria.

Pre-mortem. Think about how you are most likely to be wrong. Imagine that you are writing a letter to your future self that you will open and read once the outcome is known. In the letter you try to explain to your future self the most likely way that your forecast will be deemed to be a poor forecast. Are you most worried about missing a key piece of information? What is the biggest uncertainty and would keep you awake at night?

Don't anchor to round numbers. It's ok for probabilities to have odd numbers.

Pay close attention to the exact wording and resolution source in the resolution criteria.

You should reason both in probability space (e.g., "an event has a 5% chance of happening") as well as in odds space (e.g. "an event has a 1 in 20 chance of happening"). This helps with gaining an intuitive and self-consistent sense of the probabilities involved.

When reasoning about probabilities, think about them in the context of bets. Would you be willing to pay 90 cents for a coupon that would be out $1 if the event happens? Conversely, would you prefer to pay 10 cents to take the other side of the bet? Only if you're indifferent between the two sides have you arrived at a consistent probability.
</forecasting tips>

Before answering, use your own internal thinking extensively. Then, write a brief summary which includes at least the following items:

(a) The time left until the outcome to the question is known.
(b) The status quo outcome if nothing changed.
(c) Think about answering the question with different scopes to help ensure that you have a self consistent view and have considered the broader context.
    For example, imagine the question is: Will a company declare bankruptcy in the next 3 months? It can be useful to force yourself to forecast the probability of bankruptcy over the next 1 year, 2 years, and 5 years. Doing this in a self-consistent way helps to force you to consider the scope explicitly.
    If you forecast 40% in 12 months, then you might forecast 10% in 3 months to be scope sensitive. However, it is also possible that the bankruptcy risk is higher in the near term, so it could still be 25%.
    You will need to use your judgment. Being scope insensitive is a common cognitive bias and this exercise is meant to help combat this bias by forcing you to explicitly consider the question's scope.
(d) A brief reasoning for why a No outcome might be likely.
(e) A brief reasoning for why a Yes outcome.
(f) Reason about your probabilities in terms of bets and make sure that you'd be indifferent between both sides.

The last thing you write is your final answer. This should be the probability you assign to a "YES" outcome, written as a percentage, i.e., a number between 0 and 100.
\end{lstlisting}
\end{tcolorbox}

\subsection{Deduplication prompt}
\label{sec:appendix-deduplication-prompt}

The following prompt was used for the question deduplication process.

\begin{tcolorbox}[
    breakable,
    width=\textwidth,
    left=10pt,
    right=10pt,
    top=6pt,
    bottom=6pt,
    boxrule=0.5pt,
    colback=gray!10,
    colframe=gray!50,
    title=Deduplication Prompt
]
\begin{lstlisting}[
    breaklines=true,
    breakatwhitespace=true,
    basicstyle=\small\ttfamily,
    frame=single,
    framesep=2mm,
    xleftmargin=2mm,
    xrightmargin=2mm,
    backgroundcolor=\color{gray!10},
    showstringspaces=false,
    columns=flexible,
    literate={https://}{https://\allowbreak}1
            {http://}{http://\allowbreak}1
            {www.}{www.\allowbreak}1
            {.com}{.com\allowbreak}1
            {.org}{.org\allowbreak}1
            {.net}{.net\allowbreak}1
            {.edu}{.edu\allowbreak}1
            {.gov}{.gov\allowbreak}1
            {/}{/\allowbreak}1
            {?}{?\allowbreak}1
            {&}{&\allowbreak}1
            {=}{=\allowbreak}1
            {-}{-\allowbreak}1
]
Analyze if these two questions are duplicates (asking the same thing in different ways).

<question1>
<question>
Will the Joe Rogan Experience be ranked 1st on the Spotify Podcast Charts on March 31, 2025?
</question>
<description>
[American Best-Selling Author Beats Joe Rogan's $250M Spotify Podcast for a 4th Consecutive Top Spot](https://www.essentiallysports.com/ufc-mma-news-american-
best-selling-author-beats-joe-rogans-two-fifty-million-dollar-
spotify-podcast-for-a-fourth-consecutive-top-spot/)

Ensure settings are on United States and Top Podcasts
</description>
<resolution_criteria>
This question resolves as YES if The Joe Rogan Experience is ranked exactly 1st on the Spotify Podcast Charts at [this link](https://podcastcharts.byspotify.com/) when checked by Metaculus on or after March 31, 2025, and NO otherwise.
</resolution_criteria>

<question2>
<question>
Will the Joe Rogan Experience be ranked 1st on the Spotify Podcast Charts on March 31, 2025?
</question>
<description>
[American Best-Selling Author Beats Joe Rogan's $250M Spotify Podcast for a 4th Consecutive Top Spot](https://www.essentiallysports.com/ufc-mma-news-american-
best-selling-author-beats-joe-rogans-two-fifty-million-dollar-
spotify-podcast-for-a-fourth-consecutive-top-spot/)

Ensure settings are on United States and Top Podcasts
</description>
<resolution_criteria>
This question resolves as YES if The Joe Rogan Experience is ranked exactly 1st on the Spotify Podcast Charts at [this link](https://podcastcharts.byspotify.com/) when checked by Metaculus on or after March 31, 2025, and NO otherwise.
</resolution_criteria>
\end{lstlisting}
\end{tcolorbox}

\subsection{Question resolution prompt}
\label{sec:appendix-question-resolution-prompt}

The following prompt was used (with slight modifications) to resolve questions. The question, description, resolution criteria, and additional information were added in JSON format at the end of the prompt.

\begin{tcolorbox}[
    breakable,
    width=\textwidth,
    left=10pt,
    right=10pt,
    top=6pt,
    bottom=6pt,
    boxrule=0.5pt,
    colback=gray!10,
    colframe=gray!50,
    title=Question Resolution Prompt
]
\begin{lstlisting}[
    breaklines=true,
    breakatwhitespace=true,
    basicstyle=\small\ttfamily,
    frame=single,
    framesep=2mm,
    xleftmargin=2mm,
    xrightmargin=2mm,
    backgroundcolor=\color{gray!10},
    showstringspaces=false,
    columns=flexible,
    literate={https://}{https://\allowbreak}1
            {http://}{http://\allowbreak}1
            {www.}{www.\allowbreak}1
            {.com}{.com\allowbreak}1
            {.org}{.org\allowbreak}1
            {.net}{.net\allowbreak}1
            {.edu}{.edu\allowbreak}1
            {.gov}{.gov\allowbreak}1
            {/}{/\allowbreak}1
            {?}{?\allowbreak}1
            {&}{&\allowbreak}1
            {=}{=\allowbreak}1
]
I need you to to resolve a binary Metaculus question from 2025-10-09 to YES or NO.

<instructions>
<resolution>
`resolution` should be True if the question resolved YES or `False` if the question resolved NO.
Pay special attention to the resolution criteria; if in doubt, they always supersede "reasonable interpretations of the question". Be extremely literal in their interpretation!

Try extremely hard to find the resolution source as specified in the resolution criteria and don't just stop when you found some snippet suggesting a particular resolution.
</resolution>

<resolution_derivation>
This should be a bullet-proof argument for why your resolution is the correct one, that should convince even someone who just lost a bunch of money betting on this question and tries to find loopholes to overturn its resolution (either by making it resolve the other way or by having it annulled). To make this bullet-proof, include all links and all Google queries you used, so others can verify your approach!

<important>Ultimately, you will be scored not just on how accurate your resolutions are, but also---in the case of having gotten some wrong---how easy it was to spot the mistake. So make it clear when you are making assumptions, just like in the above example!</important>
<important>Absolutely make sure to mention pieces of evidence that you found and how you found them. In particular, mention links where you found information and mention ALL the Google searches used to find things _verbatim_---particularly if you are compiling a list/dataset or looking for the existence of something.</important>
</resolution_derivation>

<resolution_weaknesses>
Remember, you will be scored not just on how accurate your resolutions are, but also---in the case of having gotten some wrong---how easy it was to spot the mistake. This is the place to draw attention to potentially flawed steps in your derivation, causing issues with your resolution (e.g. how an assumption used to derive the result could fail to be true) and/or to reasons this question cannot be definitively resolved (i.e. has to be annulled).
For the sake of filling out this field, consider the hypothetical where there was ONE subtle mistake/misunderstanding you made in your derivation.
</resolution_weaknesses>
</instructions>
\end{lstlisting}
\end{tcolorbox}

\end{document}